\documentclass{article}

\usepackage[preprint]{corl_2026} 

\usepackage{graphicx}
\usepackage{amsmath}
\usepackage{amssymb}
\usepackage{hyperref}

\usepackage{booktabs}
\usepackage{diagbox}
\usepackage{wrapfig}
\usepackage{multirow}
\usepackage{makecell}
\usepackage{float}
\usepackage{array}
\usepackage[svgnames]{xcolor}
\usepackage{enumitem}

\renewcommand{\arraystretch}{0.95}

\definecolor{DarkGreen}{rgb}{0.0,0.39,0.0}

\title{WT-UMI: Tactile-based Whole-Body Manipulation \\via Force-Supervised Contact-Aware Planning}

\author{
\normalfont\small Jaehwi Jang$^{*}$, Zhaoyuan Gu$^{*}$, Alfred Cueva, Zimeng Chai, Junjie Sheng, Thong Nguyen\\
\normalfont\small Himank Galundia, Yifan Wu, Huishu Xue, Isaac Legene, Ojas Mediratta, Davin Doan\\
\normalfont\small Andrew Collins, Sarah Sadegh, KyoungMok Kim, Rishita Dhalbisoi, Zun Chen, Ye Zhao \\
\normalfont\footnotesize $^{*}$equally contributed.
}

\begin{document}
\maketitle
\begingroup
\renewcommand{\thefootnote}{}
\makeatletter
\renewcommand{\@makefntext}[1]{\parindent 0pt\noindent #1}
\makeatother
\footnotetext{\vspace{1em}\footnotesize The authors are with The Institute for Robotics and Intelligent Machines, Georgia Institute of Technology. }
\endgroup
\vspace{-1.5em}

%===============================================================================

\begin{abstract}
Whole-body humanoid manipulation of bulky, deformable, and shared-load objects requires distributed contact sensing and explicit force regulation, yet most imitation policies treat contact force only implicitly. On the other hand, different demonstration sources provide complementary modalities with inherent trade-offs: human demonstrations capture natural contact forces but not robot-executable actions, while teleoperation directly records robot actions but with less natural force regulation. This paper presents \textbf{WT-UMI}, a wearable whole-body tactile interface worn by human operators or mounted on humanoids, providing accurate observations of tactile images, contact forces, and end-effector poses across both human demonstration and humanoid teleoperation modes. We introduce a force-conditioned target-pose correction module that converts measured human poses into contact-aware robot targets by learning corrections from teleoperation data. To leverage the natural force interaction in human data, we propose a force-supervised planner that predicts end-effector pose chunks and contact-force trajectories. The predicted contact force serves as the reference for a tactile-based admittance controller. Across five contact-rich tasks spanning deformable objects, bulky rigid objects, and human--humanoid collaboration, WT-UMI improves success rate and reduces contact-position tracking error over four policy baselines. Our project page is available at \url{https://wt-umi.github.io/WTUMI/}. 
\end{abstract}

\keywords{Humanoid Whole-body Manipulation, Tactile Sensing, Robot Learning, Force-aware Planning} 

\begin{figure}[H]
    \centering
    \includegraphics[width=\linewidth]{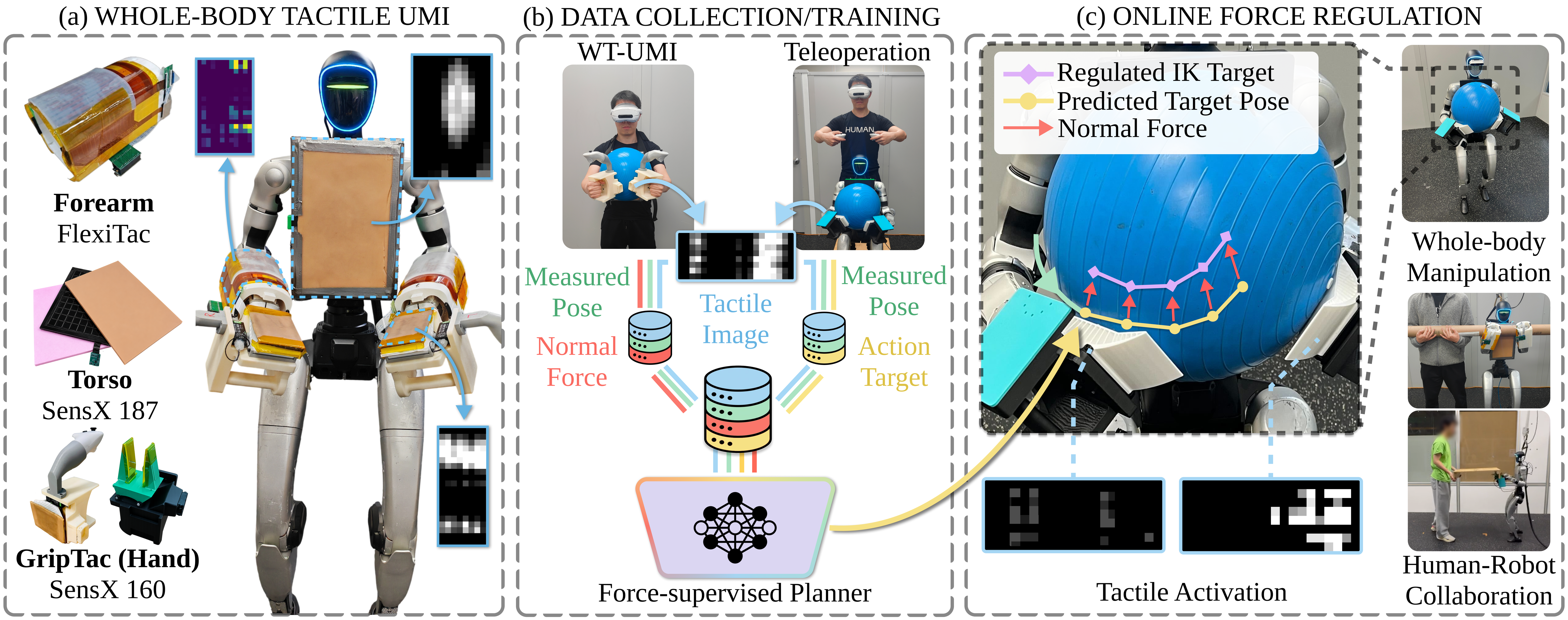}
    \caption{(a) \textbf{WT-UMI} is a shared interface between human demonstrators and humanoid robots for whole-body tactile data collection. (b) A human demonstrator wears WT-UMI, or a humanoid is controlled via teleoperation using the same hardware. (c) A force-supervised planner trained from WT-UMI data executes contact-rich tactile-aware tasks, spanning whole-body manipulation of deformable and large rigid objects and human--humanoid collaborative transport.}
    \label{fig:teaser}
\end{figure}
%===============================================================================
\section{Introduction}

%\vspace{-4mm}

Humanoid robots are increasingly expected to manipulate bulky, deformable, and shared-load objects in human environments. Tasks such as carrying a large box, reorienting a soft pillow, or transporting a beam with a human partner cannot rely on grasping alone; instead, they require coordinated contact across the torso, forearms, and hands to distribute interaction forces~\citep{gu_humanoid_2025}. Small errors in contact location or force allocation, however, can lead to slips, collisions, or load loss. Reliable whole-body manipulation therefore requires distributed contact sensing and joint motion-force planning that actively regulates both contact location and interaction force.

Most learning-based manipulation systems rely on vision and proprioception~\citep{chi2023diffusion}. However, vision is often occluded during contact and does not directly measure interaction forces, while proprioception cannot localize body-surface contact. Prior tactile sensing has focused primarily on fingertip arrays~\citep{niu2026learningversatilehumanoidmanipulation, hou2025adaptive, han2024learning, 11246486, helmut2025tactileconditioneddiffusionpolicyforceaware, zhang2026tacvlacontactawaretactilefusion}, which do not capture the distributed contact central to whole-body interaction. Existing body-mounted tactile systems~\citep{Mittendorfer_tactile2013, Punyo_SR} often rely on model-based controllers or per-task reward shaping, limiting their ability to learn diverse skills. Turning distributed whole-body tactile sensing into a contact force plan for humanoid whole-body manipulation remains an open problem.

We address this problem with \textbf{WT-UMI} (\textbf{W}hole-Body \textbf{T}actile \textbf{U}niversal \textbf{M}anipulation \textbf{I}nterface), a wearable tactile interface paired with a \textbf{force-aware} learning framework for humanoid whole-body manipulation (Fig.~\ref{fig:teaser}).
WT-UMI enables scalable human demonstration collection while reducing the human--humanoid embodiment mismatch. 
The human demonstrations collected through WT-UMI capture whole-body, contact-rich interactions with calibrated contact forces, enabling explicit contact-force prediction during training and force regulation during deployment. However, human demonstrations often lack robot-executable action labels. To convert these force-aware human demonstrations into robot-executable actions, we introduce a \textbf{force-conditioned target-pose correction} module that learns target-pose corrections from robot-in-the-loop teleoperation. Human and teleoperation trajectories are paired using force-inferred contact modes, with teleoperation commands supervising the correction of measured human poses. The resulting corrected target poses serve as action label to supervise planner training. The action-labeled human data preserves accurate contact-force measurements, enabling a \textbf{force-supervised planner} to jointly predict contact forces and corrected actions (Fig.~\ref{fig:framework}). During deployment, a tactile-based admittance controller then tracks the predicted force trajectory by modulating the predicted end-effector poses, achieving stable and force-regulated contact. 

Overall, our framework offers four key contributions.
(i) We introduce \textbf{WT-UMI}, a wearable whole-body tactile interface that collects tactile images and force-supervised demonstrations. The same sensing hardware supports both human demonstrations and humanoid teleoperation.
(ii) We design a \textbf{force-conditioned target-pose correction} module that converts human trajectories into contact-aware robot actions by learning pose corrections from teleoperation data.
(iii) We propose a \textbf{force-supervised planner} whose cross-attention head predicts the normal contact-force trajectory, with the predicted force serving as the reference for a tactile-based admittance controller.
(iv) We validate the framework on five contact-rich whole-body tasks spanning deformable objects, bulky rigid objects, and human--humanoid collaboration, where it improves the success rate and reduces contact-position drift over four policy baselines.

\section{Related Work} 

\paragraph{Whole-Body Tactile Sensing and Manipulation.}
Whole-body tactile sensing is an emerging modality for contact-rich manipulation, yet its adoption remains challenging~\citep{gu_humanoid_2025}. Much prior work restricts contact to the hands, leaving the torso and forearms largely uncovered~\citep{cheng2026tacumimultimodaluniversalmanipulation}. Compliant-body designs such as Punyo~\citep{Punyo_SR} extend coverage with pressure-sensitive skins on the arms and chest, while discrete-cell skins such as HEX-o-SKIN~\citep{Mittendorfer_tactile2013} cover larger areas but require per-cell kinematic calibration. In contrast, our thin-film piezoresistive arrays produce dense 2D contact images with large-area coverage that integrate naturally with vision encoders and imitation-learning pipelines. Existing whole-body manipulation methods span model-based control~\citep{Armleder_Gordon_huamnoid_2025, Murooka_2024}, planning~\citep{Subburaman_plan_estimate_tactile_humanoid_2025}, reinforcement learning~\citep{zheng2025embracingbulkyobjectshumanoid, liu2024opt2skill, chip, HMC, softmimic, wu2025learn, refinedp}, and imitation learning~\citep{chi2023diffusion, Murooka_TACT}. However, these methods typically rely on accurate contact models, object-state estimation, and task-specific reward design, rather than learning an explicit contact-force plan for whole-body humanoid manipulation.

\paragraph{Joint Motion-Force Prediction.}
Model-based hybrid motion-force control requires explicit force sensing and careful contact modeling~\citep{ Armleder_Gordon_huamnoid_2025, Khatib_IJRR_OSC, wijayarathne2023real}. Learning-based approaches such as ForceMimic and UMI-FT learn motion-force representations from force-motion capture or wrist/finger force sensing, but their measurements remain local~\citep{ForceMimic_2025, umift}. Recent tactile- and force-aware policies further show the value of physical grounding for contact-rich manipulation, through tactile-conditioned force actions, tactile-force representation learning, force-feedback fusion, or force-token distillation~\citep{helmut2025tactileconditioneddiffusionpolicyforceaware, huang2026tafvlatactileforcealignmentvisionlanguageaction,NEURIPS2025_8633b46e, li2026forcevla2unleashinghybridforceposition, zhao2026fdvlaforcedistilledvisionlanguageactionmodel}. However, these methods primarily focus on local fingertip, gripper, and wrist contacts for tabletop tasks. Humanoid Touch Dreaming~\citep{niu2026learningversatilehumanoidmanipulation} predicts future tactile signals in a learned latent space, but the predictions act as auxiliary regularization and contact is modeled only implicitly. 
Our method instead (i) predicts contact force explicitly from distributed tactile arrays spanning the robot body, and (ii) uses the predicted force as a reference for a tactile-based admittance controller, yielding explicit force-aware control during whole-body manipulation.

\paragraph{Demonstration Interfaces for Behavior Cloning.}
Teleoperation provides direct robot action labels, with VR, retargeting, and recent portable systems improving ergonomics and coverage~\citep{zhang2018deep, qin_anyteleop_2023, mobile_aloha, aldaco2024aloha, ze2026twist2}. However, it still requires robot access, task setup, and skilled operators. In contrast, UMI-style interfaces bypass robot-in-the-loop collection by recording in-the-wild human demonstrations with handheld devices and transferring them to robot policies~\citep{chi2024umi}. Existing interfaces mainly provide kinematic supervision and visual observations, without distributed whole-body contact force for explicit force planning. When touch is included, human-to-robot transfer becomes more challenging since human and robot embodiments, as well as their tactile sensors, may differ. TactAlign tackles this issue by aligning human and robot tactile observations in a shared latent space using rectified flow, without paired data or manual labels~\citep{wi2026tactalignhumantorobotpolicytransfer}. WT-UMI takes a complementary route: it uses shared wearable/robot-mounted tactile hardware to reduce sensing mismatch, while combining human force-rich demonstrations with robot teleoperation labels for whole-body humanoid behavior cloning. 

%===============================================================================
\section{Methods}
\begin{figure}[h]
    \centering
    \includegraphics[width=0.99\linewidth]{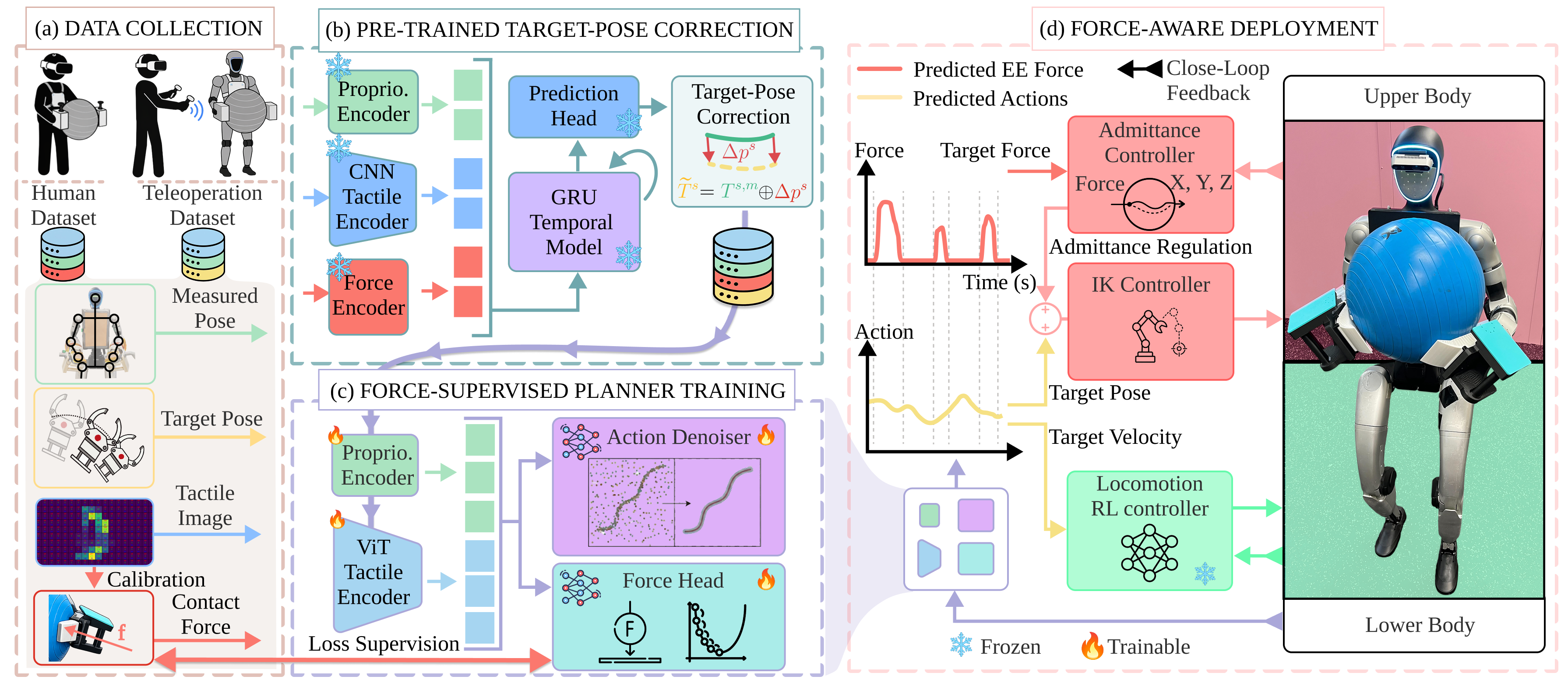}
    \caption{\small %\textbf{WT-UMI} is a wearable whole-body tactile manipulation interface that can be used by human demonstrators and deployed on a robot. 
    A force-conditioned target-pose correction module creates action labels for human data. A force-supervised planner produces a {contact-force trajectory} in addition to end-effector poses. The predicted forces are used for online force regulation via a tactile-based admittance controller.
    }
    \label{fig:framework}
    \vspace{-0.5em}
\end{figure}

Our system architecture is outlined in Fig.~\ref{fig:framework}. In \emph{data collection}, WT-UMI records distributed tactile readings and calibrated contact force from a human operator or via teleoperation on a humanoid robot. In \emph{pre-trained target-pose correction}, a force-conditioned model learns to correct human target poses into robot-executable actions. In \emph{planner training}, an action denoiser predicts corrected bimanual pose chunks, while a force head predicts contact-force trajectories. During \emph{force-aware deployment}, the predicted forces drive a tactile-based admittance controller that regulates target poses. A pre-trained RL policy simultaneously controls lower-body locomotion.

\subsection{Whole-Body Tactile Universal Manipulation Interface (WT-UMI) for Data Collection}
\begin{figure}[!ht]
    \vspace{-1.2em}
    \centering
    \includegraphics[width=0.8\columnwidth , scale=2.0]{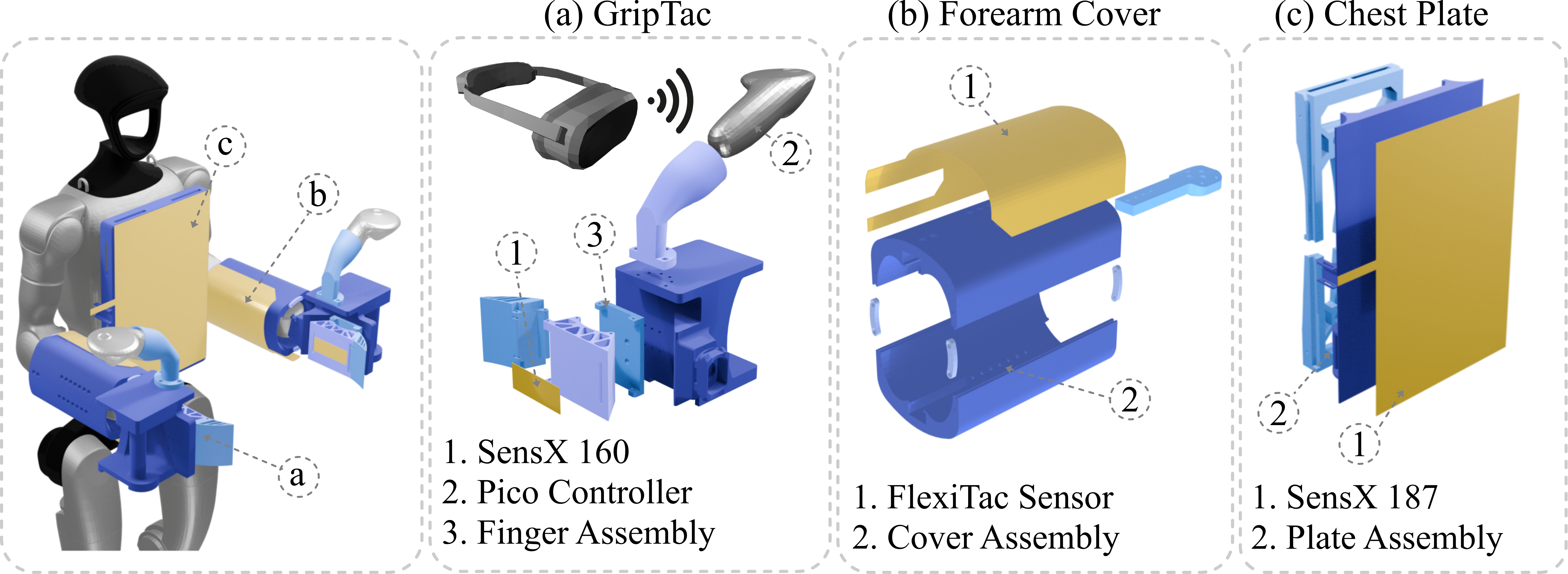}
    \caption{\small \textbf{WT-UMI} includes GripTac end-effectors, forearm covers, and a chest plate, each equipped with a thin-film tactile sensor.
    }
    \label{fig:g1_sensors}
    \vspace{-1.2em}
\end{figure}

We develop WT-UMI, an extensible wearable system for capturing operator motion and distributed tactile feedback. As shown in Fig.~\ref{fig:g1_sensors}, it integrates tactile sensors on hardware modules shared between humans and humanoids: a chest plate, forearm covers, and handheld GripTacs, which are tactile-instrumented handheld interfaces with interchangeable end-effectors. By using the same sensing modules for human data collection, robot teleoperation, and robot execution, WT-UMI reduces the domain gap between demonstration and deployment (see Fig.~\ref{fig:teaser}).

WT-UMI supports demonstration collection in two modes (Fig.~\ref{fig:framework}). In \emph{teleoperation mode}, WT-UMI is mounted on the robot, and an operator streams bimanual pose commands through a PICO VR headset~\citep{pico4ultra2023} and two handheld controllers via XRoboToolkit~\citep{11404528}. In \emph{human mode},
a human operator wears the chest plate, forearm covers, and GripTacs, with a PICO controller on each GripTac to track the bimanual pose. Human mode requires no robot in the loop. 
% , yet still captures the force interactions needed for whole-body manipulation. 
Both modes stream hand poses, tactile images, and calibrated force measurements; teleoperation additionally records robot proprioception and VR commands. More details about sensor specifications and force calibration are provided in Appendix~\ref{app:calibration}. 

\subsection{Force-Conditioned Target-Pose Correction}
\label{sec:ik_correction}
We introduce a force-conditioned target-pose correction module that converts the measured human hand pose $T^{s,\mathrm{m}}$, contact force $f^{s,\mathrm{m}}$, and tactile observation $\mathbf{I}^{s,\mathrm{m}}$ into a robot target pose, where $s\in\{l,r\}$ indexes the left and right hands and $\mathrm{m}$ denotes measured quantities. This module is pre-trained first and then applied to human demonstrations to generate robot-executable action labels to supervise planner training.
% \Yez{Is this anticipated force $f^s$? If so, be explicit and accurate. \ZGU{Updated}}
The correction module is pre-trained offline using paired robot teleoperation and human demonstration trajectories. For both data sources, we use contact forces to infer contact modes, such as right-contact, both-contact, and left-contact. %segmenting each trajectory at changes in contact state, yielding  segments.
We then match contact modes between teleoperation and human data using Dynamic Time Warping (DTW) over the end-effector motion. Given the aligned contact modes, the target-pose offset from the robot teleoperation data is used to supervise the corresponding offset in the human data, gated by inferred contact state: only hands that are in contact for that segment receive a nonzero target-pose offset.
For each hand $s$, a lightweight CNN-GRU correction network $g_\theta$ predicts the hand-specific translation offset from the measured tactile image, pose, and force, $\Delta\mathbf{p}^{s}=g_\theta(\mathbf{I}^{s,\mathrm{m}},T^{s,\mathrm{m}},f^{s,\mathrm{m}})\in\mathbb{R}^3$, over sampled aligned teleoperation--human trajectory pairs.
After pre-training, the learned correction module is applied to human demonstration data to generate target-pose action labels. The corrected target pose in SE(3) is $\tilde{T}^{s}=T^{s,\mathrm{m}}\oplus\mathbf{1}[f^{s,\mathrm{m}}\ge f_{\text{threshold}}]\Delta\mathbf{p}^{s}$, where $\oplus$ shifts the translation component. The resulting corrected poses $\tilde{T}^{s}$ serve as robot-executable action labels for training an action denoiser.
The model is trained with a Smooth-$L_1$ offset loss and a temporal smoothness penalty; additional training details are provided in Appendix~\ref{app:modulation}.
    \vspace{-0.5em}
\subsection{Force-Supervised Planner}
\label{sec:force_supervision}
Existing generative planners often do not explicitly model contact forces, leaving force regulation to position-based feedback alone. WT-UMI addresses this limitation by providing calibrated normal force from tactile measurements. We leverage these calibrated forces in the force-conditioned target-pose correction module and use them as supervision to augment the planner with a force head that predicts future contact-force trajectories. During deployment, the predicted forces serve as explicit references for a tactile-based admittance controller. 

The planner operates at $50$ Hz with $400$ ms observation and prediction horizons ($H_o = H_a = 20$). At timestep $t$, the model consumes an observation history $\mathbf{o}_{t-H_o+1:t}$, where each $\mathbf{o}_t = (\mathbf{I}^{l}_t,\; \mathbf{I}^{r}_t,\; \mathbf{I}^{ch}_t,\; \mathbf{T}_t)$ consists of normalized tactile images from the left/right contact sites and the chest, alongside the measured bimanual end-effector pose $\mathbf{T}_t = [\mathrm{vec}(T^{l}_t);\, \mathrm{vec}(T^{r}_t)]\in \mathbb{R}^{18}$, where $\mathrm{vec}(\cdot)$ maps each SE(3) pose $T^{s}_t \in \mathbb{R}^{4\times4}$, $s\in\{l,r\}$, to its 3D translation and 6D continuous rotation representation~\citep{zhou2019continuity} in $\mathbb{R}^{9}$. A vision transformer (ViT)~\citep{VIT} encodes the channel-stacked tactile images and a two-layer MLP encodes the hand poses; per timestep, {their outputs are concatenated} into one observation token of width $D = 320$. The hand pose is additionally linearly projected to a separate pose token of the same width, yielding a conditioning sequence $\mathbf{h}\in\mathbb{R}^{2H_o\times D}$. Conditioned on $\mathbf{h}$, a transformer-based action denoising policy (either flow matching~\citep{lipman2023flow} or diffusion~\citep{chi2023diffusion}) generates a predicted action chunk $\mathbf{a} = [\mathbf{T}_{t+1},\dots,\mathbf{T}_{t+H_a}] \in \mathbb{R}^{H_a \times 18}$. The action denoising policy is supervised by the target-pose-corrected labels $\mathbf{a}^{\star}=[\tilde{\mathbf{T}}_{t+1},\dots,\tilde{\mathbf{T}}_{t+H_a}]\in \mathbb{R}^{H_a \times 18}$ from Sec.~\ref{sec:ik_correction}.

Unlike the action denoising policy, a force head performs direct regression without a diffusion process. It consists of a cross-attention decoder that takes $H_a$ learnable positional queries $\mathbf{Q} \in \mathbb{R}^{H_a \times D}$, whose $t$-th row $\mathbf{q}_t \in \mathbb{R}^{D}$ is a positional query embedding for predicting the force at the $t$-th future action step, where $D = 320$ matches the encoder token width, and uses the shared observation embeddings $\mathbf{h}$ as keys and values.  By performing self-attention across the queries followed by cross-attention to $\mathbf{h}$, the decoder models the temporal force dynamics before a linear projection maps the output to the predicted force chunk of both hands $\mathbf{F} \in \mathbb{R}^{H_a \times 2}$. Supervision is provided by the calibrated ground-truth forces $\mathbf{F}^{\star} \in \mathbb{R}^{H_a \times 2}$, with a softplus projection enforcing physical non-negativity. Because the force head directly outputs a force trajectory, its gradients flow back through $\mathbf{h}$, allowing the shared encoder to be jointly optimized by both the action and force objectives. To suppress gradient spikes from contact onset, the force head is supervised via an element-wise Huber loss ($\mathrm{SmoothL1}$), yielding the combined objective:
$
\mathcal{L} = \mathcal{L}_{\text{gen}}(\mathbf{a}, \mathbf{a}^{\star}) + \lambda_F\,\mathrm{SmoothL1}\!\left(\mathbf{F},\mathbf{F}^{\star}\right),
$
where $\mathcal{L}_{\text{gen}}$ is a flow-matching or diffusion denoising MSE and $\lambda_F$ is the weight balancing the action denoiser and force head losses. More training details can be found in Appendix~\ref{sec:hyperparam}.

\subsection{Tactile-based Admittance Controller}
\label{sec:lowlevel}
The low-level controller is split into a lower body ($12$ leg joints and $3$ waist joints) and an upper body ($14$ arm joints). The lower body tracks pelvis-frame velocity $\mathbf{v} = [v_x, v_y, \omega_z]^\top$ with a pre-trained RL locomotion policy~\citep{gr00tn1_2025} that maintains balance under arbitrary upper-body motion. The upper body is regulated by a tactile-based admittance controller. At each control timestep $t$, we track the predicted end-effector pose target $\tilde{T}^{s}_{t}$ and a normal-force reference $f^{s}_{t}$ for hand $s$. In addition to the learned target-pose correction, a proportional admittance controller regulates the motion and force simultaneously. This controller refines the planner-generated pose using both normal-force and contact-centroid feedback:
$
T^{s}_{\text{cmd},t}
= \tilde{T}^{s}_{t}\,\Delta T^{s}_t, 
$
where $T^{s}_{\text{cmd},t}$ is the commanded target pose. The admittance regulation $\Delta T^{s}_t \in \mathbb{R}^{4 \times 4}$ is a corrective SE(3) increment for palm $s$ expressed in the palm frame, incorporating the corrective rotation $\Delta R^{s}_{xy,t}$ and translation $\Delta\mathbf{p}^{s}_{z,t}$ derived from proportional admittance law based on force-measurement feedback.  The commanded poses are then passed to an optimization-based inverse kinematics solver~\citep{pink}, $q_{\text{cmd},t} = \mathrm{IK}(T^{l}_{\text{cmd},t},\,T^{r}_{\text{cmd},t},\,q_t,\,\dot{q}_t)$.  Joint-level tracking uses a PD controller with gravity compensation, $\tau_t = K_p(q_{\text{cmd},t} - q_t) + K_d(-\dot{q}_t) + G(q_t)$, where $q_t, \dot{q}_t$ are measured joint positions and velocities, $q_{\text{cmd},t}$ is the IK solution, $\tau_t$ is the commanded joint torque, $K_p, K_d$ are joint-PD gains, and $G(q_t)$ is the gravity-compensation torque.

\section{Experiment Setup}

\paragraph{Tasks.}
\label{sec:tasks}
We demonstrate five contact-rich tasks in three categories, with representative deployments shown in Fig.~\ref{fig:deployment}. For \textbf{deformable object manipulation}, \emph{yoga ball manipulation} (\textbf{T1}) stabilizes and repositions an inflated ball, and \emph{pillow reorientation} (\textbf{T2}) reorients a soft pillow. For \textbf{bulky rigid object manipulation}, \emph{bucket manipulation} (\textbf{T3}) repositions a cone-shaped container with diverse loads. For \textbf{human--humanoid collaborative manipulation},
\emph{beam transport} (\textbf{T4}) and \emph{table transport} (\textbf{T5}) require the robot and a human partner to jointly carry a beam and a table, respectively, while inferring human physical intent from tactile feedback and following the partner's motion. 

\paragraph{Policy Baselines.}
We train four widely used behavior-cloning policies as baselines: \emph{ViT-FMT}~\citep{lipman2023flow}, \emph{ViT-DiT}~\citep{chi2023diffusion}, $\pi_{0.5}$~\citep{pi05}, and $\Psi_0$~\citep{wei2026psi0openfoundationmodel}. 
\emph{ViT-FMT} and \emph{ViT-DiT} share the same vision transformer encoder but differ in the generative process: flow matching for \emph{ViT-FMT} and denoising diffusion for \emph{ViT-DiT}. For ViT-FMT and ViT-DiT, the ViT tactile encoder is trained end-to-end with both heads. $\pi_{0.5}$ and $\Psi_0$ are fine-tuned foundation policies: $\pi_{0.5}$ is a vision-language-action model conditioned on tactile images and a fixed per-task language instruction; we adopt the OpenPI-based implementation in FASTER~\citep{lu2026fasterrethinkingrealtimeflow}.  $\Psi_0$ is a humanoid foundation policy conditioned on tactile images and proprioception. For both foundation policies, encoders are frozen during fine-tuning. 

\paragraph{Hardware Setup.}
We use a Unitree G1 humanoid for teleoperation data collection and policy deployment. Raw tactile data streams at $100$\,Hz and proprioception streams at $500$\,Hz; both are resampled to a synchronized $50$\,Hz stream during training and deployment. The ViT-FMT and ViT-DiT policies run on an RTX~4500 and perform asynchronous inference with a $100$\,ms chunk latency. $\pi_{0.5}$ and $\Psi_0$ similarly run at $12$\,Hz with $83$\,ms per chunk on an RTX~6000. The low-level tactile-based admittance controller runs at $200$\,Hz. For ViT-DiT, we use DDIM~\citep{song2021denoising} for faster inference. All policies adopt training-time real-time chunking (RTC)~\citep{black2025trainingtimeactionconditioningefficient} to compensate for inference delay. 

\section{Results}
\label{sec:results}
\begin{figure}[t]
    \centering
    \includegraphics[width=0.99\columnwidth]{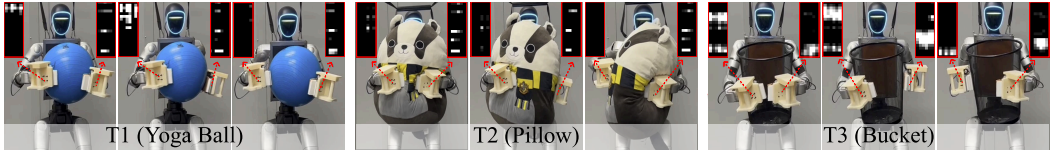} 
    \caption{\small Deployment of our framework on three whole-body manipulation tasks. }
    \label{fig:deployment}
    \vspace{-0.5em}
\end{figure}

We evaluate WT-UMI across five tasks introduced in Sec.~\ref{sec:tasks}.  Representative deployments for T1--T3 are shown in Fig.~\ref{fig:deployment} and used for quantitative evaluation throughout Sec.~\ref{sec:results}. For the collaborative tasks T4 and T5, we detail their setup and successful deployments in Appendix~\ref{sec:transport}.

\subsection{Force Head Evaluation}
\label{sec:results_force_head}
This section investigates whether the force head produces accurate, smooth, and temporally aligned force predictions. To address this, we evaluate our force head (in Sec.~\ref{sec:force_supervision}) within a \textit{ViT-FMT} policy. We employ a 90/10 train-validation split across both teleoperation and human datasets derived from task T1.
We evaluate the trained force-supervised planner on $10$ held-out demonstrations and compare predicted forces with ground-truth forces.

\begin{wraptable}{r}{0.6\columnwidth}
  \vspace{-1.3em}
  \centering
  \footnotesize
  \setlength{\tabcolsep}{3pt}
  \begin{tabular}{l|cccc}
  \toprule
  \textbf{Source} & \textbf{Force RMSE (N)} & \textbf{Lag (ms)} & \multicolumn{2}{c}{\textbf{Force Rate RMS}} \textbf{(N/s)}\\[2pt]
  \cline{4-5}
  \noalign{\vspace{2pt}}
   &     &   & \textbf{Meas.} & \textbf{Pred.} \\
  \midrule
  Human  & \textbf{1.05} & \textbf{68}  & \textbf{5.86}  & \textbf{3.74}  \\
  Teleop & 2.07          & 151          & 30.62          & 19.80          \\
  \bottomrule
  \end{tabular}
  \vspace{0.0em}
  \caption{Force signal quality. Human demonstrations exhibit more accurate and smoother force profiles than teleoperation, improving the quality of
  supervision for learning.}
  \label{tab:data_force_quality}
  \vspace{-1em}
  \end{wraptable}

Table~\ref{tab:data_force_quality} shows the force prediction accuracy using RMSE, smoothness using force rate $|\mathrm{d}F/\mathrm{d}t|$ RMS averaged across episodes, and temporal alignment using the cross-correlation peak offset between predicted and ground-truth forces. Overall, the force head tracks the ground-truth force trajectory with low prediction error and small temporal lag. 
Notably, the force head trained on human demonstrations achieves an RMSE of $1.05$~N out of a $5.56$~N force RMS and a $2.2\times$ lower temporal lag. This improved temporal consistency is further reflected in the lower force-rate RMS, which decreases to $3.74$~N/s compared to the $19.80$~N/s from the force head trained on teleoperation data. %\Yez{is 3.74 an absolute small value already? I can not tell that, although it is much smaller than 19.8.}
  
\subsection{Effect of Training Data Sources and Force-Conditioned Target-Pose Correction}
\label{sec:results_data}
This section evaluates the complementary roles of teleoperated robot data and human demonstrations collected with WT-UMI. We evaluate ViT-FMT because it is the best-performing policy among all baselines, as demonstrated in Sec.~\ref{sec:result_policy}. Among the representative tasks \textbf{T1}--\textbf{T3}, T2 involves a soft pillow whose compliance keeps contact-force readings small in magnitude, and T3's bucket requires precise contact on its cone-shaped surfaces, making both tasks challenging. 
We adopt two data-source settings: pure teleoperation data (\textbf{Tel.}) or a combined dataset (\textbf{Comb.}) that augments teleoperation with target-pose-corrected human demonstrations. Across T1--T3, \textbf{Tel.} contains 2.2, 2.2, and 1.4\,min, respectively, while \textbf{Comb.} contains 13.2, 15.7, and 8.8\,min. Thus, teleoperation accounts for only 16.7\%, 14.0\%, and 15.9\% of the corresponding \textbf{Comb.} demonstrations. 
Each setting is evaluated with $N=25$ trials per task. In each trial, the object is rotated once, starting and ending in the same hugging pose. We report the following metrics: (i) success rate; (ii) contact off-center drift, defined as the distance between the measured contact centroid and the geometric center of the tactile sensor, which indicates how well the contact position remains centered; (iii) mean contact force, which measures contact firmness during required contact phases; and (iv) translational and rotational accelerations of end-effector poses, which assess motion smoothness.

\begin{table*}[h]
\centering
\small
\setlength{\tabcolsep}{3pt}
\renewcommand{\arraystretch}{0.92}
\caption{Effect of data source over tasks \textbf{T1}-\textbf{T3} (Sec.~\ref{sec:tasks}). Best values are highlighted in bold.}
\label{tab:data_source}
\begin{tabular*}{\textwidth}{@{\extracolsep{\fill}} l | cc cc cc cc cc @{}}
\toprule
\textbf{Task}
& \multicolumn{2}{c}{\textbf{Succ. (\%)}}
& \multicolumn{2}{c}{\textbf{Cont. Drift (mm)}}
& \multicolumn{2}{c}{\textbf{Cont. Force (N)}}
& \multicolumn{2}{c}{\textbf{Smooth.-Trans. ($\text{m/s}^2$)}}
& \multicolumn{2}{c}{\textbf{Smooth.-Rot. ($\text{rad/s}^2$)}}\\
\cmidrule(lr){2-3}\cmidrule(lr){4-5}\cmidrule(lr){6-7}\cmidrule(lr){8-9}\cmidrule(lr){10-11}
\multicolumn{1}{l}{\textbf{Data~$\rightarrow$}\hspace{-0.6em}} & Tel. & Comb. & Tel. & Comb. & Tel. & Comb. & Tel. & Comb. & Tel. & Comb. \\
\midrule
T1 & \textbf{100.0} & \textbf{100.0} & 20.41 & \textbf{18.12} & 3.27 & \textbf{4.77} & 4.08 & \textbf{3.14} & 22.93 & \textbf{20.29} \\
T2 & \textbf{100.0} & \textbf{100.0} & 24.61 & \textbf{21.04} & 0.17 & \textbf{0.52} & 6.03 & \textbf{1.87} & 34.54 & \textbf{12.63} \\
T3 & 60.0  & \textbf{80.0}        & 26.01 & \textbf{25.00} & 0.66 & \textbf{0.96} & 3.49 & \textbf{1.85} & 27.82 & \textbf{14.05} \\
\bottomrule
\end{tabular*}
\end{table*}

Table~\ref{tab:data_source} presents the quantitative results. The \textbf{Tel.} policy alone achieves a strong $86.7\%$ success rate on average, because teleoperation provides robot-feasible action labels. However, teleoperation lacks accurate force feedback during data collection, reflected in its occasional missed contacts or over-pressing behaviors, which result in larger contact-region drift of about $11\%$ on average compared to \textbf{Comb.}. Conversely, human demonstrations offer valuable contact information because the demonstrator directly perceives interaction forces and naturally regulates contact. However, pure human data is not directly robot-executable due to missing action labels and the human--humanoid kinematics gap, which leads to poor policy performance when used alone. Our proposed human-data correction addresses this limitation by converting human data into robot-feasible pose targets. In the \textbf{Comb.} policy, the corrected human data supplements teleoperation data by reducing contact center drift and increasing mean contact force by approximately $2$ times on average.
Motion smoothness also improves consistently across all tasks with the \textbf{Comb.} policy: translational and rotational accelerations are reduced by $46.3\%$ and $41.5\%$ on average, respectively. The \textbf{Comb.} policy also increases the success rate by 20\% for T3, where the most common failure mode is loss of contact and motion freeze due to out-of-distribution observations. Notably, the target-pose correction module requires only a small amount of teleoperation data to convert a much larger set of human demonstrations into robot-executable data, demonstrating the teleoperation-data efficiency of the proposed module. 

\subsection{Effect of Force-Conditioned Admittance Control}
\label{sec:results_ik_admittance}

This section evaluates whether tactile-based admittance control improves contact quality and motion stability.
We evaluate ViT-FMT across tasks \textbf{T1}--\textbf{T3}, all trained on \textbf{Combined} human and teleoperation datasets. The result is shown in Table~\ref{tab:ablation_admittance_policies}. Across tasks \textbf{T1}--\textbf{T3}, enabling admittance control consistently improves motion smoothness, reducing translational acceleration by $10.4\%$ and rotational acceleration by $9.0\%$ on average. Contact quality also improves: contact off-center drift decreases by $10.9\%$ and mean contact force increases by $2.7\%$ on average, indicating more stable and better-centered contact when contact is desired. The overall result confirms that our admittance controller substantially improves motion smoothness while maintaining stable, firm contact.

\begin{table*}[t]
\centering
\small
\setlength{\tabcolsep}{4pt}
\renewcommand{\arraystretch}{0.92}
\caption{Admittance ablation across policy backbones over tasks \textbf{T1}, \textbf{T2}, and \textbf{T3}. Each metric is evaluated without (w/o) and with (w/) our admittance control. Best or tied values are bolded.}
\label{tab:ablation_admittance_policies}
\begin{tabular}{ll | cc  cc  cc  cc @{}}
\toprule
\textbf{Policy}
& \textbf{Task}
& \multicolumn{2}{c|}{\textbf{Cont. Drift (mm)}}
& \multicolumn{2}{c|}{\textbf{Cont. Force (N)}}
& \multicolumn{2}{c|}{\textbf{Smooth.-Trans. ($\text{m/s}^2$)}}
& \multicolumn{2}{c}{\textbf{Smooth.-Rot. ($\text{rad/s}^2$)}} \\
\cmidrule(lr){3-4}\cmidrule(lr){5-6}\cmidrule(lr){7-8}\cmidrule(lr){9-10}
\multicolumn{2}{c}{\textbf{Admi. (Ours) $\rightarrow$}} & w/o & w/ & w/o & w/ & w/o & w/ & w/o & w/ \\
\midrule
\multirow{3}{*}{ViT-FMT}
& T1  & 18.12 & \textbf{15.67} & 4.77 & \textbf{5.50} & 3.14 & \textbf{2.98} & 20.29 & \textbf{18.56} \\
& T2  & 21.04 & \textbf{19.44} & \textbf{0.52} & 0.13 & \textbf{1.87} & 1.95 & \textbf{12.63} & 13.61 \\
& T3  & 25.00 & \textbf{22.08} & 0.96 & \textbf{1.61} & 1.85 & \textbf{1.29} & 14.05 & \textbf{10.38} \\
\bottomrule
\end{tabular}
\end{table*}

\subsection{Effect of Policy Backbone on Whole-body Manipulation Tasks}
\label{sec:result_policy}
In this section, we compare the baseline policies $\pi_{0.5}$, $\Psi_0$, ViT-DiT, and ViT-FMT. All policies are trained on the same combined dataset with force prediction and deployed with the admittance controller (deployment videos in the supplementary material). Overall, ViT-FMT achieves the best motion smoothness, as reflected by the lowest translational and rotational accelerations ($2.07~\text{m/s}^2$ and $14.18~\text{rad/s}^2$, respectively), and most closely reproduces the motion patterns in the dataset. $\pi_{0.5}$ ($4.62~\text{m/s}^2$, $28.45~\text{rad/s}^2$) shows more jittery motion, often with rapid end-effector swings. The ViT-DiT ($2.90~\text{m/s}^2$, $18.35~\text{rad/s}^2$) and $\Psi_0$ ($3.38~\text{m/s}^2$, $22.83~\text{rad/s}^2$) tend to stuck in static poses or fail to continue the motion, leading to frequent start-stop behavior and larger accelerations.

%===============================================================================
\section{Conclusion}

In this study, we presented WT-UMI, a whole-body humanoid manipulation system built together with a force-conditioned target-pose correction module and a force-supervised planner that leverage force-rich human demonstrations. The correction module converts human hand poses into contact-aware robot target poses by applying learned corrections from teleoperated robot data, yielding action labels for planner training. The force-supervised planner uses a cross-attention force head to predict a contact-force trajectory via direct regression. At deployment, a tactile-based admittance controller consumes the predicted force as the normal-force reference, maintaining stable contact. Together, the target-pose correction, force-supervised planner, human-robot co-training data, and admittance controller improve contact-rich whole-body manipulation across four policy backbones, spanning deformable, large rigid, and human-collaborative tasks.

\section{Limitations and Future Work}

Our system has three main limitations. First, tactile coverage is constrained by available sensor configurations, so only the palms, forearms, and chest are instrumented. Extending coverage to dexterous hands, legs, and back would broaden the set of tactile-driven tasks WT-UMI can support. Second, our policy does not yet consume RGB vision input, because third-person views introduce a human--humanoid embodiment gap and finding a camera angle that consistently avoids object occlusion is challenging, especially for transport tasks. Fusing tactile sensing with vision is an important next step toward enabling the policy to anticipate future contact and re-establish lost contact. 
Third, the force head predicts only a scalar normal force; extending it to multi-axis contact wrenches and distributed force maps would support finer force regulation in more complex contact configurations.
%===============================================================================

\clearpage
\acknowledgments{The authors thank the members of The Institute for Robotics and Intelligent Machines at Georgia Institute of Technology for their support and feedback. The authors also thank Dr. Yunzhu Li, Binghao Huang, and their team for FlexiTac sensor support.}

%===============================================================================

\bibliography{references}

@misc{niu2026learningversatilehumanoidmanipulation,
      title={Learning Versatile Humanoid Manipulation with Touch Dreaming}, 
      author={Yaru Niu and Zhenlong Fang and Binghong Chen and Shuai Zhou and Revanth Senthilkumaran and Hao Zhang and Bingqing Chen and Chen Qiu and H. Eric Tseng and Jonathan Francis and Ding Zhao},
      year={2026},
      eprint={2604.13015},
      archivePrefix={arXiv},
      primaryClass={cs.RO},
}

@article{han2024learning,
  title={Learning generalizable vision-tactile robotic grasping strategy for deformable objects via transformer},
  author={Han, Yunhai and Yu, Kelin and Batra, Rahul and Boyd, Nathan and Mehta, Chaitanya and Zhao, Tuo and She, Yu and Hutchinson, Seth and Zhao, Ye},
  journal={IEEE/ASME Transactions on Mechatronics},
  volume={30},
  number={1},
  pages={554--566},
  year={2024},
  publisher={IEEE}
}

@article{chi2023diffusion,
author = {Cheng Chi and Zhenjia Xu and Siyuan Feng and Eric Cousineau and Yilun Du and Benjamin Burchfiel and Russ Tedrake and Shuran Song},
title ={Diffusion policy: Visuomotor policy learning via action diffusion},

journal = {The International Journal of Robotics Research},
volume = {44},
number = {10-11},
pages = {1684-1704},
year = {2025},
eprint = { 
    
        https://doi.org/10.1177/02783649241273668
    
    

}
}

@article{wijayarathne2023real,
  title={Real-time deformable-contact-aware model predictive control for force-modulated manipulation},
  author={Wijayarathne, Lasitha and Zhou, Ziyi and Zhao, Ye and Hammond, Frank L},
  journal={IEEE Transactions on Robotics},
  volume={39},
  number={5},
  pages={3549--3566},
  year={2023},
  publisher={IEEE}
}

@inproceedings{wu2025learn,
  author    = {Wu, Feiyang and Nal, Xavier and Jang, Jaehwi and Zhu, Wei and Gu, Zhaoyuan and Wu, Anqi and Zhao, Ye},
  title     = {Learn to Teach: Sample-Efficient Privileged Learning for Humanoid Locomotion over Real-World Uneven Terrain},
  booktitle = {IEEE Robotics and Automation Letters},
  year      = {2025},
  pages     = {9048--9055}
}

@InProceedings{mobile_aloha,
  title = 	 {Mobile ALOHA: Learning Bimanual Mobile Manipulation using Low-Cost Whole-Body Teleoperation},
  author =       {Fu, Zipeng and Zhao, Tony Z. and Finn, Chelsea},
  booktitle = 	 {Proceedings of The 8th Conference on Robot Learning},
  pages = 	 {4066--4083},
  year = 	 {2025},
  volume = 	 {270}
}

@inproceedings{chi2024umi,
  title={Universal Manipulation Interface: In-The-Wild Robot Teaching Without In-The-Wild Robots},
  author={Chi, Cheng and Xu, Zhenjia and Pan, Chuer and Cousineau, Eric  and Burchfiel, Benjamin and Feng, Siyuan and Tedrake, Russ and Song, Shuran},
  booktitle={Proceedings of Robotics: Science and Systems},
  pages={p045},
  year={2024}
}

@article{Punyo_SR,
author = {Jose A. Barreiros  and Aykut Özgün Önol  and Mengchao Zhang  and Sam Creasey  and Aimee Goncalves  and Andrew Beaulieu  and Aditya Bhat  and Kate M. Tsui  and Alex Alspach },
title = {Learning contact-rich whole-body manipulation with example-guided reinforcement learning},
journal = {Science Robotics},
volume = {10},
number = {105},
pages = {eads6790},
year = {2025},
eprint = {https://www.science.org/doi/pdf/10.1126/scirobotics.ads6790}}

@article{aldaco2024aloha,
  title={ALOHA 2: An Enhanced Low-Cost Hardware for Bimanual Teleoperation},
  author={Aldaco, Jorge and Armstrong, Travis and Baruch, Robert and Bingham, Jeff and Chan, Sanky and Draper, Kenneth and Dwibedi, Debidatta and Finn, Chelsea and Florence, Pete and Goodrich, Spencer and others},
  year={2024}
}

@article{Khatib_IJRR_OSC,
author = {Oussama Khatib and Mikael Jorda and Jaeheung Park and Luis Sentis and Shu-Yun Chung},
title ={Constraint-consistent task-oriented whole-body robot formulation: Task, posture, constraints, multiple contacts, and balance},

journal = {The International Journal of Robotics Research},
volume = {41},
number = {13-14},
pages = {1079-1098},
year = {2022},
eprint = { 
    
        https://doi.org/10.1177/02783649221120029
}
}

@inproceedings{zhang2018deep,
  title={Deep imitation learning for complex manipulation tasks from virtual reality teleoperation},
  author={Zhang, Tianhao and McCarthy, Zoe and Jow, Owen and Lee, Dennis and Chen, Xi and Goldberg, Ken and Abbeel, Pieter},
  booktitle={IEEE International Conference on Robotics and Automation},
  pages={5628--5635},
  year={2018}
}

@inproceedings{qin_anyteleop_2023,
	title = {{AnyTeleop}: {A} {General} {Vision}-{Based} {Dexterous} {Robot} {Arm}-{Hand} {Teleoperation} {System}},
	isbn = {978-0-9923747-9-2},
	shorttitle = {{AnyTeleop}},
	language = {en},
	urldate = {2024-08-16},
	booktitle = {Robotics: {Science} and {Systems}},
	author = {Qin, Yuzhe and Yang, Wei and Huang, Binghao and Wyk, Karl and Su, Hao and Wang, Xiaolong and Chao, Yu-Wei and Fox, Dieter},
	year = {2023},
}

@ARTICLE{liu2024opt2skill,
  author={Liu, Fukang and Gu, Zhaoyuan and Cai, Yilin and Zhou, Ziyi and Jung, Hyunyoung and Jang, Jaehwi and Zhao, Shijie and Ha, Sehoon and Chen, Yue and Xu, Danfei and Zhao, Ye},
  journal={IEEE Robotics and Automation Letters}, 
  title={Opt2Skill: Imitating Dynamically-Feasible Whole-Body Trajectories for Versatile Humanoid Loco-Manipulation}, 
  year={2025},
  volume={10},
  number={11},
  pages={12261-12268}
  }

@ARTICLE{gu_humanoid_2025,
  author={Gu, Zhaoyuan and Li, Junheng and Shen, Wenlan and Yu, Wenhao and Xie, Zhaoming and McCrory, Stephen and Cheng, Xianyi and Shamsah, Abdulaziz and Griffin, Robert and Liu, C. Karen and Kheddar, Abderrahmane and Peng, Xue Bin and Zhu, Yuke and Shi, Guanya and Nguyen, Quan and Cheng, Gordon and Gao, Huijun and Zhao, Ye},
  journal={IEEE/ASME Transactions on Mechatronics}, 
  title={Humanoid Locomotion and Manipulation: Current Progress and Challenges in Control, Planning, and Learning}, 
  year={2026},
  volume={31},
  number={2},
  pages={2300-2330}}

@inproceedings{ze2026twist2,
  title={TWIST2: Scalable, Portable, and Holistic Humanoid Data Collection System},
  author={Ze, Yanjie and Zhao, Siheng and Wang, Weizhuo and Kanazawa, Angjoo and Duan, Rocky and Abbeel, Pieter and Shi, Guanya and Wu, Jiajun and Liu, C. Karen},
  booktitle={IEEE International Conference on Robotics and Automation},
  year={2026}
}

@inproceedings{gr00tn1_2025,
  archivePrefix = {arxiv},
  eprint     = {2503.14734},
  title      = {{GR00T} {N1}: An Open Foundation Model for Generalist Humanoid Robots},
  author     = {NVIDIA and Johan Bjorck and Fernando Castañeda, Nikita Cherniadev and Xingye Da and Runyu Ding and Linxi "Jim" Fan and Yu Fang and Dieter Fox and Fengyuan Hu and Spencer Huang and Joel Jang and Zhenyu Jiang and Jan Kautz and Kaushil Kundalia and Lawrence Lao and Zhiqi Li and Zongyu Lin and Kevin Lin and Guilin Liu and Edith Llontop and Loic Magne and Ajay Mandlekar and Avnish Narayan and Soroush Nasiriany and Scott Reed and You Liang Tan and Guanzhi Wang and Zu Wang and Jing Wang and Qi Wang and Jiannan Xiang and Yuqi Xie and Yinzhen Xu and Zhenjia Xu and Seonghyeon Ye and Zhiding Yu and Ao Zhang and Hao Zhang and Yizhou Zhao and Ruijie Zheng and Yuke Zhu},
  year       = {2025},
}

@ARTICLE{Murooka_TACT,
  author={Murooka, Masaki and Hoshi, Takahiro and Fukumitsu, Kensuke and Masuda, Shimpei and Hamze, Marwan and Sasaki, Tomoya and Morisawa, Mitsuharu and Yoshida, Eiichi},
  journal={IEEE Robotics and Automation Letters}, 
  title={TACT: Humanoid Whole-Body Contact Manipulation Through Deep Imitation Learning With Tactile Modality}, 
  year={2025},
  volume={10},
  number={8},
  pages={7819-7826}
  }

@INPROCEEDINGS{Mittendorfer_tactile2013,
  author={Mittendorfer, Philipp and Yoshida, Eichii and Moulard, Thomas and Cheng, Gordon},
  booktitle={IEEE/RSJ International Conference on Intelligent Robots and Systems}, 
  title={A general tactile approach for grasping unknown objects with a humanoid robot}, 
  year={2013},
  volume={},
  number={},
  pages={4747-4752}
  }

@inproceedings{zheng2025embracingbulkyobjectshumanoid,
  title={Embracing bulky objects with humanoid robots: Whole-body manipulation with reinforcement learning},
  author={Zheng, C. and Chen, K. and Bi, Z. and Li, Y. and Pan, L. and Zhou, J. and Li, H. and Ma, J.},
  booktitle={IEEE International Conference on Robotics and Automation},
  pages={16930},
  year={2026}
}

@article{Armleder_Gordon_huamnoid_2025,
author = {Armleder, Simon and Bergner, Florian and Guadarrama-Olvera, Julio Rogelio and Nakanishi, Jun and Cheng, Gordon},
title = {Real-Time Control of a Humanoid Robot for Whole-Body Tactile Interaction},
journal = {Advanced Intelligent Systems},
volume = {7},
number = {12},
pages = {e202500149},
eprint = {https://advanced.onlinelibrary.wiley.com/doi/pdf/10.1002/aisy.202500149},
year = {2025}
}

@INPROCEEDINGS{Subburaman_plan_estimate_tactile_humanoid_2025,
  author={Subburaman, Rajesh and Stasse, Olivier},
  booktitle={IEEE-RAS International Conference on Humanoid Robots}, 
  title={A Whole-Body Multi Contact Large Object Manipulation and Estimation Framework for Humanoids Using Skin Patches}, 
  year={2025},
  volume={},
  number={},
  pages={1-8},
  }

@ARTICLE{Murooka_2024,
  author={Murooka, Masaki and Fukumitsu, Kensuke and Hamze, Marwan and Morisawa, Mitsuharu and Kaminaga, Hiroshi and Kanehiro, Fumio and Yoshida, Eiichi},
  journal={IEEE Robotics and Automation Letters}, 
  title={Whole-Body Multi-Contact Motion Control for Humanoid Robots Based on Distributed Tactile Sensors}, 
  year={2024},
  volume={9},
  number={11},
  pages={10620-10627}
  }

@INPROCEEDINGS{ForceMimic_2025,
  author={Liu, Wenhai and Wang, Junbo and Wang, Yiming and Wang, Weiming and Lu, Cewu},
  booktitle={IEEE International Conference on Robotics and Automation}, 
  title={ForceMimic: Force-Centric Imitation Learning with Force-Motion Capture System for Contact-Rich Manipulation}, 
  year={2025},
  volume={},
  number={},
  pages={1105-1112},
}

@misc{black2025trainingtimeactionconditioningefficient,
      title={Training-Time Action Conditioning for Efficient Real-Time Chunking}, 
      author={Kevin Black and Allen Z. Ren and Michael Equi and Sergey Levine},
      year={2025},
      eprint={2512.05964},
      archivePrefix={arXiv},
      primaryClass={cs.RO},
}

@software{pink,
  title = {{Pink: Python inverse kinematics based on Pinocchio}},
  author = {Caron, Stéphane and De Mont-Marin, Yann and Budhiraja, Rohan and Bang, Seung Hyeon and Domrachev, Ivan and Nedelchev, Simeon and Du, Peter and Escande, Adrien and Vaillant, Joris and Wingo, Bruce and Patapati, Santosh and San José Pro, Daniel},
  license = {Apache-2.0},
  version = {4.1.0},
  year = {2026}
}

@inproceedings{HMC,
  title={HMC: Learning Heterogeneous Meta-Control for Contact-Rich Loco-Manipulation},
  author={Wei, Lai and Peng, Xuanbin and Qiu, Ri-Zhao and Huang, Tianshu and Cheng, Xuxin and Wang, Xiaolong},
  booktitle={IEEE International Conference on Robotics and Automation},
  year={2026}
}

@inproceedings{
VIT,
title={An Image is Worth 16x16 Words: Transformers for Image Recognition at Scale},
author={Alexey Dosovitskiy and Lucas Beyer and Alexander Kolesnikov and Dirk Weissenborn and Xiaohua Zhai and Thomas Unterthiner and Mostafa Dehghani and Matthias Minderer and Georg Heigold and Sylvain Gelly and Jakob Uszkoreit and Neil Houlsby},
booktitle={International Conference on Learning Representations},
year={2021}
}

@inproceedings{umift,
      title={In-the-Wild Compliant Manipulation with UMI-FT}, 
      author={Hojung Choi and Yifan Hou and Chuer Pan and Seongheon Hong and Austin Patel and Xiaomeng Xu and Mark R. Cutkosky and Shuran Song},
      year={2026},
      booktitle={IEEE International Conference on Robotics and Automation}
}

@article{chip,
    title={CHIP: Learning Adaptive Compliance for Humanoid Control through Hindsight Perturbation},
    author={Chen, Sirui and Cao, Zi-ang and Luo, Zhengyi and Castañeda, Fernando and Li, Chenran and Wang, Tingwu and Yuan, Ye and Fan, Linxi and Liu, C Karen and Zhu, Yuke},
    year={2025}
}

@article{softmimic,
      title={{SoftMimic}: Learning Compliant Whole-body Control from Examples},
      author={Margolis, Gabriel B. and Wang, Michelle and Fey, Nolan and Agrawal, Pulkit},
      year={2025}
  }

@misc{touchtronix_sensx,
  author       = {{TouchTronix Robotics Inc.}},
  title        = {{SensX thin-film tactile sensors}},
  howpublished = {{https://www.touchtronix.io/}}
}

@inproceedings{hou2025adaptive,
                title={Adaptive compliance policy: Learning approximate compliance for diffusion guided control},
                author={Hou, Yifan and Liu, Zeyi and Chi, Cheng and Cousineau, Eric and Kuppuswamy, Naveen and Feng, Siyuan and Burchfiel, Benjamin and Song, Shuran},
                booktitle={IEEE International Conference on Robotics and Automation},
                pages={4829--4836},
                year={2025}
                }

@inproceedings{
song2021denoising,
title={Denoising Diffusion Implicit Models},
author={Jiaming Song and Chenlin Meng and Stefano Ermon},
booktitle={International Conference on Learning Representations},
year={2021}
}

@misc{pico4ultra2023,
  author       = {{PICO Immersive Pte. Ltd.}},
  title        = {{PICO 4 Ultra: An All-New Mixed Reality Experience}},
  year         = {2023},
  howpublished = {\url{https://www.picoxr.com/global/products/pico4-ultra}}
}

@INPROCEEDINGS{11404528,
  author={Zhao, Zhigen and Yu, Liuchuan and Jing, Ke and Yang, Ning},
  booktitle={IEEE/SICE International Symposium on System Integration}, 
  title={{{XRoboToolkit}}: A Cross-Platform Framework for Robot Teleoperation}, 
  year={2026},
  volume={},
  number={},
  pages={15-20},
}

@InProceedings{pi05,
  title = {$\pi_{0.5}$: a Vision-Language-Action Model with Open-World Generalization},
  author =       {Kevin Black and Noah Brown and James Darpinian and Karan Dhabalia and Danny Driess and Adnan Esmail and Michael Equi and Chelsea Finn and Niccolo Fusai and Manuel Y. Galliker and Dibya Ghosh and Lachy Groom and Karol Hausman and Brian Ichter and Szymon Jakubczak and Tim Jones and Liyiming Ke and Devin LeBlanc and Sergey Levine and Adrian Li-Bell and Mohith Mothukuri and Suraj Nair and Karl Pertsch and Allen Z. Ren and Lucy Xiaoyang Shi and Laura Smith and Jost Tobias Springenberg and Kyle Stachowicz and James Tanner and Quan Vuong and Homer Walke and Anna Walling and Haohuan Wang and Lili Yu and Ury Zhilinsky},
  booktitle = 	 {Proceedings of The 9th Conference on Robot Learning},
  pages = 	 {17--40},
  year = 	 {2025},
  volume = 	 {305}
}

@inproceedings{wei2026psi0openfoundationmodel,
  author    = {Wei, Songlin and Jing, Hongyi and Li, Boqian and Zhao, Zhenyu and Mao, Jiageng and Ni, Zhenhao and He, Sicheng and Liu, Jie and Liu, Xiawei and Kang, Kaidi and Zang, Sheng and Yuan, Weiduo and Pavone, Marco and Huang, Di and Wang, Yue},
  title     = {$\Psi_0$: An Open Foundation Model Towards Universal Humanoid Loco-Manipulation},
  booktitle = {Proceedings of Robotics: Science and Systems},
  year      = {2026},
}

@inproceedings{vitac,
  title = {{3D-ViTac}: Learning Fine-Grained Manipulation with Visuo-Tactile Sensing},
  author = {Huang, Binghao and Wang, Yixuan and Yang, Xinyi and Luo, Yiyue and Li, Yunzhu},
  booktitle = {Proceedings of The 8th Conference on Robot Learning},
  pages = {2557--2578},
  year = {2025},
  volume = {270},
  eprint = {2410.24091},
  archivePrefix = {arXiv},
  primaryClass = {cs.RO}
}

@inproceedings{cheng2026tacumimultimodaluniversalmanipulation,
  author    = {Cheng, Tailai and Chen, Kejia and Chen, Lingyun and Zhang, Liding and Zhang, Yue and Ling, Yao and Hamad, Mahdi and Bing, Zhenshan and Wu, Fan and Sharma, Karan and Knoll, Alois},
  title     = {Tac{UMI}: A Multi-Modal Universal Manipulation Interface for Contact-Rich Tasks},
  booktitle = {Extended Abstracts of the ACM/IEEE International Conference on Human-Robot Interaction},
  pages     = {342--343},
  year      = {2026}
}

@misc{zhang2026tacvlacontactawaretactilefusion,
      title={TacVLA: Contact-Aware Tactile Fusion for Robust Vision-Language-Action Manipulation}, 
      author={Kaidi Zhang and Heng Zhang and Zhengtong Xu and Zhiyuan Zhang and Md Rakibul Islam Prince and Xiang Li and Xiaojing Han and Yuhao Zhou and Arash Ajoudani and Yu She},
      year={2026},
      eprint={2603.12665},
      archivePrefix={arXiv},
      primaryClass={cs.RO}
}

@misc{lu2026fasterrethinkingrealtimeflow,
      title={FASTER: Rethinking Real-Time Flow VLAs}, 
      author={Yuxiang Lu and Zhe Liu and Xianzhe Fan and Zhenya Yang and Jinghua Hou and Junyi Li and Kaixin Ding and Hengshuang Zhao},
      year={2026},
      eprint={2603.19199},
      archivePrefix={arXiv},
      primaryClass={cs.RO}
}

@inproceedings{helmut2025tactileconditioneddiffusionpolicyforceaware,
  author =		 "Helmut, E. and  Funk, N. and  Schneider, T. and  de Farias, C. and  Peters, J.",
  year =		 "2026",
  title =		 "Tactile-Conditioned Diffusion Policy for Force-Aware Robotic Manipulation",
  booktitle =		 "IEEE International Conference on Robotics and Automation",
}

@INPROCEEDINGS{11246486,
  author={Helmut, Erik and Dziarski, Luca and Funk, Niklas and Belousov, Boris and Peters, Jan},
  booktitle={IEEE/RSJ International Conference on Intelligent Robots and Systems }, 
  title={Learning Force Distribution Estimation for the GelSight Mini Optical Tactile Sensor Based on Finite Element Analysis}, 
  year={2025},
  volume={},
  number={},
  pages={8553-8560}
  }

@inproceedings{lipman2023flow,
  title={Flow Matching for Generative Modeling},
  author={Lipman, Yaron and Chen, Ricky T. Q. and Ben-Hamu, Heli and Nickel, Maximilian and Le, Matthew},
  booktitle={International Conference on Learning Representations},
  year={2023}
}

@inproceedings{NEURIPS2025_8633b46e,
 author = {Yu, Jiawen and Liu, Hairuo and Yu, Qiaojun and Ren, Jieji and Hao, Ce and Ding, Haitong and Huang, Guangyu and Huang, Guofan and Song, Yan and Cai, Panpan and Zhang, Wenqiang and Lu, Cewu},
 booktitle = {Advances in Neural Information Processing Systems},
 pages = {93409--93439},
 title = {ForceVLA: Enhancing VLA Models with a Force-aware MoE for Contact-rich Manipulation},
 volume = {38},
 year = {2025}
}

@misc{wi2026tactalignhumantorobotpolicytransfer,
      title={TactAlign: Human-to-Robot Policy Transfer via Tactile Alignment}, 
      author={Youngsun Wi and Jessica Yin and Elvis Xiang and Akash Sharma and Jitendra Malik and Mustafa Mukadam and Nima Fazeli and Tess Hellebrekers},
      year={2026},
      eprint={2602.13579},
      archivePrefix={arXiv},
      primaryClass={cs.RO}
}

@misc{huang2026tafvlatactileforcealignmentvisionlanguageaction,
      title={TaF-VLA: Tactile-Force Alignment in Vision-Language-Action Models for Force-aware Manipulation}, 
      author={Yuzhe Huang and Pei Lin and Wanlin Li and Daohan Li and Jiajun Li and Jiaming Jiang and Chenxi Xiao and Ziyuan Jiao},
      year={2026},
      eprint={2601.20321},
      archivePrefix={arXiv},
      primaryClass={cs.RO}
}

@misc{li2026forcevla2unleashinghybridforceposition,
      title={ForceVLA2: Unleashing Hybrid Force-Position Control with Force Awareness for Contact-Rich Manipulation}, 
      author={Yang Li and Zhaxizhuoma and Hongru Jiang and Junjie Xia and Hongquan Zhang and Jinda Du and Yunsong Zhou and Jia Zeng and Ce Hao and Jieji Ren and Qiaojun Yu and Cewu Lu and Yu Qiao and Jiangmiao Pang},
      year={2026},
      eprint={2603.15169},
      archivePrefix={arXiv},
      primaryClass={cs.RO}
}

@misc{zhao2026fdvlaforcedistilledvisionlanguageactionmodel,
      title={FD-VLA: Force-Distilled Vision-Language-Action Model for Contact-Rich Manipulation},
      author={Zhao, Ruiteng and Wang, Wenshuo and Ma, Yicheng and Li, Xiaocong and Tay, Francis E. H. and Ang, Jr., Marcelo H. and Zhu, Haiyue},
      year={2026},
      eprint={2602.02142},
      archivePrefix={arXiv},
      primaryClass={cs.RO} 
}

@misc{refinedp,
      title={REFINE-DP: Diffusion Policy Fine-tuning for Humanoid Loco-manipulation via Reinforcement Learning}, 
      author={Zhaoyuan Gu and Yipu Chen and Zimeng Chai and Alfred Cueva and Thong Nguyen and Yifan Wu and Huishu Xue and Minji Kim and Isaac Legene and Fukang Liu and Matthew Kim and Ayan Barula and Yongxin Chen and Ye Zhao},
      year={2026},
      eprint={2603.13707},
      archivePrefix={arXiv},
      primaryClass={cs.RO},
}

@inproceedings{zhou2019continuity,
  title={On the Continuity of Rotation Representations in Neural Networks},
  author={Zhou, Yi and Barnes, Connelly and Lu, Jingwan and Yang, Jimei and Li, Hao},
  booktitle={the IEEE/CVF Conference on Computer Vision and Pattern Recognition},
  pages={5745--5753},
  year={2019}
}

%===============================================================================
\clearpage
\section{Supplementary}

\subsection{Sensor Specification and Force Calibration}
\label{app:calibration}
As shown in Fig.~\ref{fig:teaser}, WT-UMI uses a TouchTronix SensX 187 sensor on the chest, SensX 160 sensors on the palm end-effectors~\citep{touchtronix_sensx}, and custom $16 \times 26$ FlexiTac-style sensors on the forearms~\citep{vitac}. 

We calibrate the palm TouchTronix SensX 160 sensors under controlled quasi-static normal loading from $0$ to $25$ N. Two contact configurations are evaluated: direct loading on the bare sensor surface and loading through a compliant gel pad with a thickness of $5\,\mathrm{mm}$ and a Shore-A-20 hardness. The calibrated tactile response is used as a proxy measurement of contact normal force. In this work, only the palm sensors are quantitatively force-calibrated, while the forearm and chest sensors are primarily used for contact localization and interaction-state inference.

\begin{figure}[!ht]
    \centering
    \includegraphics[width=0.6\columnwidth]{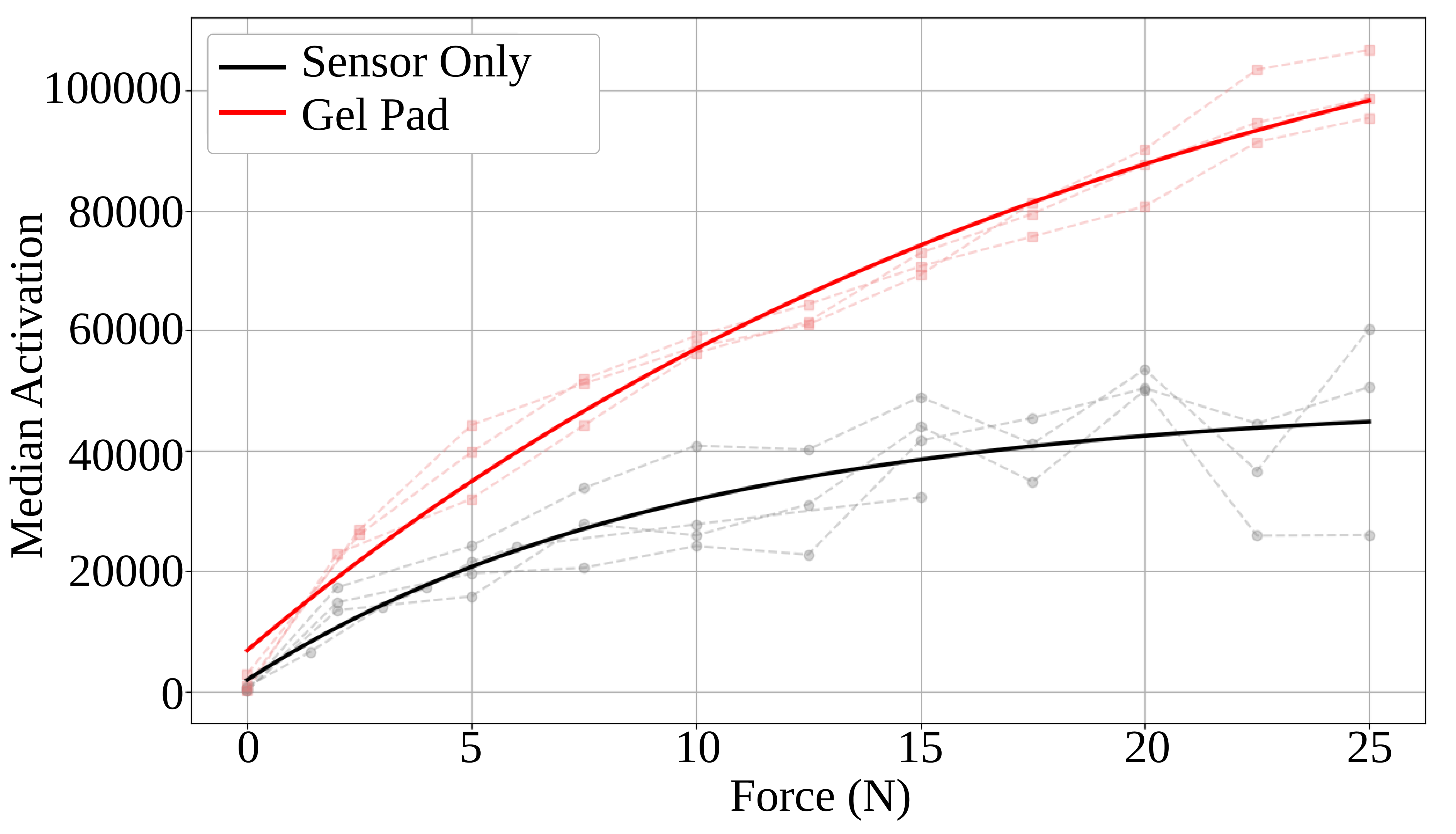}
    \caption{Force calibration of the SensX 160 palm sensor with and without a gel pad. The bare sensor response rises quickly and saturates near $20$\,N. The gel-pad configuration distributes contact pressure more evenly, yielding a smoother and less rapidly saturating response while extending the usable calibration range.}
    \label{fig:force_calibration}
\end{figure}

As shown in Fig.~\ref{fig:force_calibration}, the bare sensor produces a steep activation increase under low loads before saturating near $20\,\mathrm{N}$. In contrast, the gel-pad configuration spreads the contact load across the sensing surface, yielding a smoother and less rapidly saturating response. This improved response facilitates more stable force calibration and extends the force range before saturation. We use this gel-pad configuration for all manipulation experiments in our study. In addition to producing a more stable and consistent calibration response, the compliant gel surface increases contact friction, which improves grasp stability during contact-rich interactions while preserving a consistent force scale between WT-UMI demonstrations and robot deployment.

The contact normal force is calibrated based on the tactile activation map. For each palm sensor, each tactile frame is subtracted by the reading at zero load to be converted into a calibrated activation map. Let $X(t) \in \mathbb{R}^{H \times W}$ denote the calibrated tactile activation map at time $t$ over the taxel grid $\Omega=\{1,\ldots,H\}\times\{1,\ldots,W\}$, with $X_{ij}(t)$ representing the activation at taxel row $i$ and column $j$. Active taxels above a fixed threshold $\tau_a$ form the index set $\mathcal{I}(t)$, whose values are summed into an aggregate activation score $A(t)$. The threshold $\tau_a$ is selected empirically from no-contact recordings to suppress background sensor noise and inactive taxels, and is kept fixed across all calibration and deployment data:
\begin{equation}
    \mathcal{I}(t)
    =
    \{(i,j)\in\Omega \mid X_{ij}(t) > \tau_a\},
    \qquad
    A(t) =
    \sum_{\substack{(i,j)\in\Omega\\ X_{ij}(t)>\tau_a}}
    X_{ij}(t).
\end{equation}

The calibration fits an inverse-exponential response model between the activation score $A(t)$ and applied normal force $f(t)$. This form captures the rapidly saturating response commonly observed in piezoresistive tactile sensors under increasing normal load:
\begin{equation}
    A(t) = c_1\left(1 - \exp(-c_2 f(t))\right) + c_3,
\end{equation}
where $c_1, c_2, c_3 \in \mathbb{R}$ are scalar parameters fitted from the calibration data. Inverting this model gives an intermediate force estimate
\begin{equation}
    \tilde f(t)
    =
    -\frac{1}{c_2}
    \log\left(
        1 -
        \operatorname{clip}\left(
            \frac{A(t)-c_3}{c_1},\,0,\,1-\epsilon
        \right)
    \right),
    \label{eq:force_calibration_inverse}
\end{equation}
where $\epsilon > 0$ is a small numerical margin that keeps the argument of $\log(\cdot)$ strictly positive. The final normal-force estimate $f^{m}_{t}$ is obtained by applying a linear post-calibration scaling and offset correction to compensate for sensor-dependent gain and baseline variation. Specifically, a scale factor $\alpha$ and offset $\beta$ are fitted from the calibration measurements and applied to the intermediate force estimate, followed by clipping to the calibrated operating range $[0, f_{\max}]$,
\begin{equation}
    f^{m}_{t}
    =
    \operatorname{clip}\left(
        \alpha\,\tilde f(t) + \beta,\,
        0,\,
        f_{\max}
    \right).
\end{equation}

\subsection{Tactile-based Calibration Procedure for Human Data Collection}
Collecting human demonstrations with WT-UMI requires an online calibration procedure. The calibration process records transformations among the PICO headset, the handheld GripTacs, and the chest plate, ensuring that the recorded human motion can be consistently mapped to the corresponding robot poses.

During calibration, the operator wears the headset and chest plate while holding the GripTacs. Each GripTac's tip is pressed against the chest tactile sensor, activating exactly one cell on the tactile sensor. The resulting tactile image allows the sensor to precisely localize the contact location of the GripTac on the chest plate. Combined with the known GripTac and chest plate geometry, this measurement estimates the transformation from the chest-plate center to the handheld GripTac, denoted as $T^s_{\rm Chest\rightarrow Hand}$, with $s \in \{l, r\}$ representing the left or right hand. After calibration, the GripTac positions are expressed in the chest-plate frame.

To convert human poses into robot configurations, we use the known transformation from the robot base to the mounted chest-plate center, $T_{\rm Base\rightarrow Chest}$. The corresponding robot end-effector pose is then computed as
$T^s_{\rm Base\rightarrow Hand} = T_{\rm Base\rightarrow Chest} \cdot T^s_{\rm Chest\rightarrow Hand}$.
During data inspection, the resulting bimanual targets are passed to an inverse-kinematics solver, and poses that exceed the feasibility tolerance or risk self-collision are rejected. The retained demonstrations are then verified in simulation or on robot hardware. This calibration process ensures that WT-UMI produces consistent demonstrations.

\subsection{Target-Pose Correction Training Details}

\label{app:modulation}
\begin{figure}[h]
    \centering
    \includegraphics[width=0.95\columnwidth]{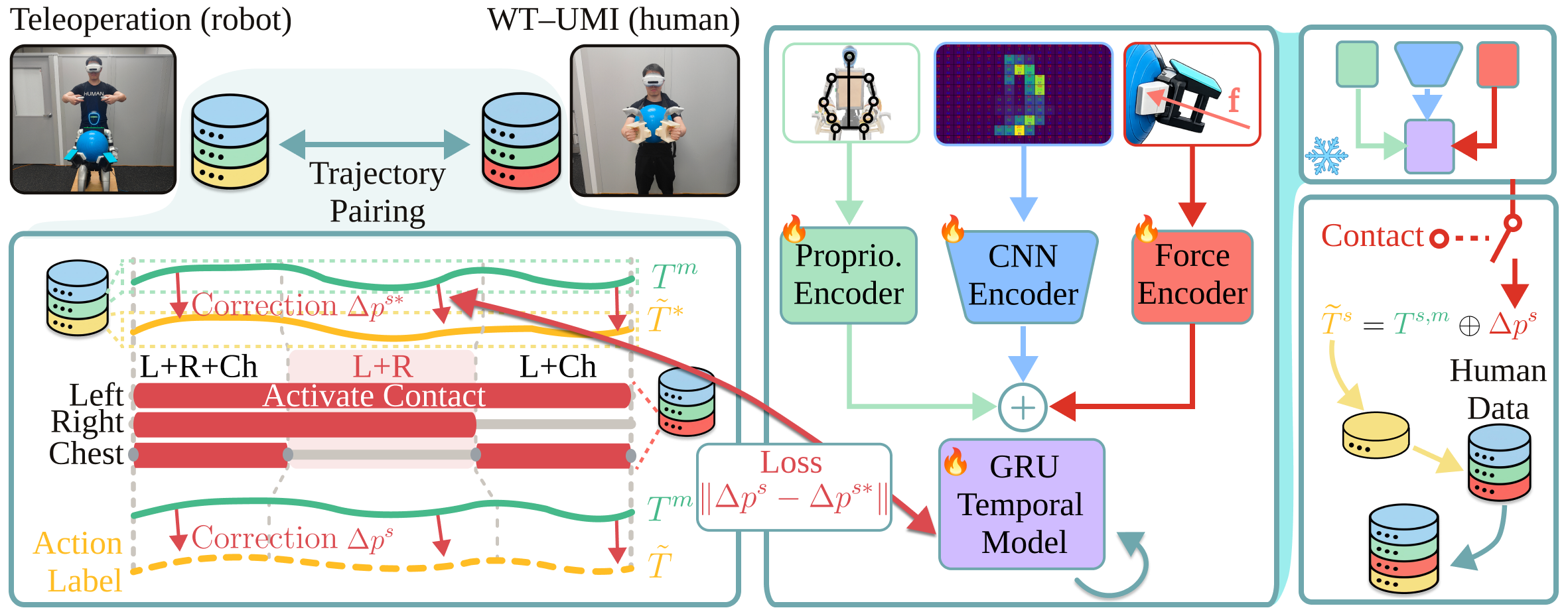}
    \caption{Target-pose correction training. Contact-mode-aligned teleoperation and human segments supervise contact-gated offsets that convert human poses into robot target poses.}
    \label{fig:ik_modulation_training}
\end{figure}

This subsection provides details of the data alignment and training procedure for the target-pose correction module in Fig.~\ref{fig:framework}(b), which produces robot-executable action labels for human data by learning contact-dependent hand-pose offsets from teleoperation data.
The module is trained separately for each task on paired teleoperation and human trajectories, as illustrated in Fig.~\ref{fig:ik_modulation_training}. Trajectories are paired by matching their contact-mode sequences (e.g., left hand + right hand + chest $\rightarrow$ left hand + right hand $\rightarrow$ left hand + chest $\rightarrow$ $\cdots)$. 
Each teleoperation trajectory supplies a residual target-pose offset $\Delta\mathbf{p}^\star$ in the local end-effector frame, that supervises the predicted correction $\Delta\mathbf{p}$ applied on top of the measured human hand pose. A deployment-time force gating ensures the correction is applied only to force-active hands.

The training data are resampled to $50$~Hz. The model observes a $0.2$~s history of hand pose, tactile images, contact force, and binary contact state for both hands, and predicts corrections over a $0.1$~s future target-pose chunk. Because the correction is applied offline to recorded demonstrations, the human hand poses in the upcoming prediction chunk are also available and provided as input. The module then predicts a per-step correction that is applied to each upcoming hand pose to obtain the robot-executable target pose.

The network combines a per-block CNN tactile encoder with a GRU temporal model: a two-layer GRU over the history and a two-layer bidirectional GRU over the future window, both with hidden dimension $256$ and dropout $0.3$. An auxiliary head predicts the contact-segment class and per-hand contact direction, while the output head regresses the translational offset $\Delta\mathbf{p}=[\Delta\mathbf{p}^l;\Delta\mathbf{p}^r]\in\mathbb{R}^{6}$, where each $\Delta\mathbf{p}^s\in\mathbb{R}^{3}$ is predicted in the local end-effector frame of hand $s\in\{l,r\}$. Training minimizes a contact-masked Smooth-$L_1$ offset loss, a Smooth-$L_1$ temporal-smoothness loss on consecutive offset differences, and the auxiliary contact losses. We optimize with AdamW using cosine learning-rate decay, warmup, and gradient clipping, with early stopping on validation offset error and auxiliary contact accuracy.

\subsection{Force Head Architecture and Hyperparameters}
\label{sec:hyperparam}

{The force head uses a TransformerDecoder with two layers and four attention heads using a hidden dimension $D=320$, with a total of 0.5M parameters. Crucially, the action denoiser and the force head are split into independent decoders that share only the encoder. The action denoiser takes the noisy action chunk $a_k$ and the denoising step $k$ as input, while the force head reads only the clean observation embedding $\mathbf{h}$. 
Keeping the decoders independent allows the force head's input to remain free of the noise schedule while still letting force supervision shape the shared encoder.}

{During training, both flow-matching and transformer backbones undergo the same number of gradient steps using AdamW, cosine learning rate decay, and an exponential moving average over weights. Image augmentations applied during training include additive noise, channel dropout, and random patch masking. The sensor cap $f_{\max}{=}20$\,N is fitted to the calibrated palm range, and the calibration-aware force term is enabled throughout training with $\lambda_F = 1.0$. At inference, the force-conditioned target-pose correction (Sec.~\ref{sec:ik_correction}) is gated by the anticipated per-hand force using $f_{\text{threshold}} = 0.5$\,N.}

\subsection{Human--Humanoid Collaborative Manipulation}

\begin{figure}[!t]
    \centering
    \includegraphics[width=0.95\columnwidth]{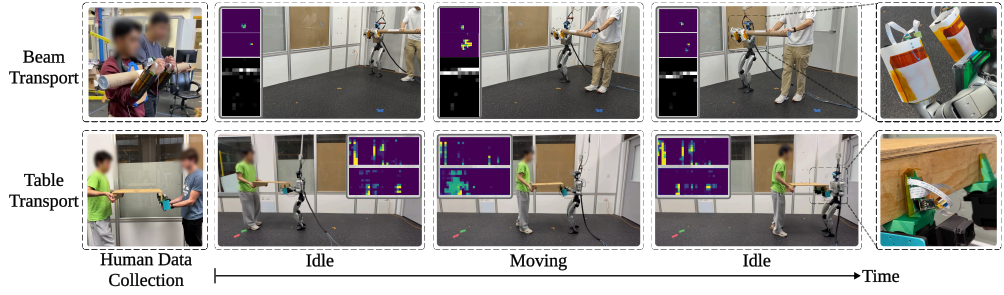}
    \caption{Human--humanoid collaborative manipulation tasks. Top images show the beam transport task, and bottom images show the table transport task. For each task, a depiction of the human-human data collection is included.}
    \label{fig:collaborative}
\end{figure}

We further evaluate WT-UMI and our planning framework on two human--humanoid collaborative manipulation tasks: beam transport (T4) and table transport (T5), shown in Fig.~\ref{fig:collaborative}. Unlike the single-agent tasks in Sec.~\ref{sec:tasks}, these tasks require the humanoid to maintain distributed whole-body contact while reading its partner's intent from tactile feedback: changes in force distribution and contact activation across the sensing surfaces signal whether the human is pushing or pulling, and the humanoid adapts its whole-body motion to accomplish coordinated transport.

\paragraph{Task Setup.}
Both tasks require continuous contact regulation and shared-load coordination. In the beam transport task, the humanoid stabilizes a cardboard beam (length $1.33$ m, diameter $0.09$ m, weight $8.72$ N) using distributed contact across the forearms and chest, while a human partner applies pushing or pulling forces to guide the motion. The robot's forearms support most of the load, while its chest provides additional stabilization and prevents slip. The tactile signals in these regions contain visible patterns for intent inference   (Fig.~\ref{fig:collaborative}): during forward motion, left forearm activation typically increases while right forearm activation decreases, and the chest activation distribution shifts with the interaction-force direction.

In the table-transport task, the humanoid supports a plywood table with both palms while adapting to translational motions initiated by a human partner. The table measures $609\,\mathrm{mm}\times609\,\mathrm{mm}\times125\,\mathrm{mm}$, with a weight of $39\,\mathrm{N}$. For this task, we use FlexiTac sensors mounted on a custom-designed gripper, as shown in Fig.~\ref{fig:teaser}.

\label{sec:transport}

\paragraph{Data Collection.} 
For both tasks, demonstrations are collected through human-human collaborative transport, where one participant wears WT-UMI and physically mimics the robot role, while the other acts as the human partner. These demonstrations capture natural shared-load coordination behaviors and tactile contact patterns without requiring robot teleoperation.  For the beam transport task, we further collect an additional robot-in-the-loop dataset to cover the embodiment mismatch between human and robot. In this data collection, WT-UMI is mounted directly on the humanoid robot. Overall, we collected $56$ beam transport trajectories and $45$ table transport trajectories containing pushing, pulling, and idle interaction behaviors, with each trajectory lasting approximately $15$ seconds on average. The action chunk in transport tasks are base velocity in SE(2): $\mathbf{a} = [\mathbf{v}_{t+1},\dots,\mathbf{v}_{t+H_a}] \in \mathbb{R}^{H_a \times 3}$.

\paragraph{Deployment.}
During deployment, the humanoid supports both forward and backward locomotion in the beam and table task, and returns to an idle state when the contact distribution relaxes to the nominal stationary-holding pattern. The transition delay is approximately $0.7$ s from idle to active transport and $0.8$ s from active transport back to idle.

\subsection{Comparison of Four Policy Backbones}

To complement the aggregated results in Sec.~\ref{sec:results_ik_admittance} and Sec.~\ref{sec:result_policy}, we compare four policy backbones ($\pi_{0.5}$, $\Psi_0$, ViT-DiT, and ViT-FMT) on tasks \textbf{T1}--\textbf{T3}, each evaluated with and without our tactile admittance controller. Each setting is evaluated with $N=25$ trials per task. Table~\ref{tab:four-policy-comparisom} reports per-task success rate together with contact-quality and motion-smoothness metrics.

\begin{table*}[h]
\centering
\small
\setlength{\tabcolsep}{4pt}
\renewcommand{\arraystretch}{0.92}
\caption{Admittance ablation across all policy backbones over tasks \textbf{T1}, \textbf{T2}, and \textbf{T3}. Each metric is evaluated without (w/o) and with (w/) our admittance control. The better or tied value is bolded for success rate, contact drift, and motion smoothness.\\\textit{Note: these appendix evaluations are obtained from separate runs from those in Sec.~\ref{sec:result_policy}; due to variations in experiments, the values may differ.}}
\label{tab:four-policy-comparisom}
\begin{tabular}{ll | cc  cc  cc  cc  cc @{}}
\toprule
\textbf{Policy}
& \textbf{Task}
& \multicolumn{2}{c|}{\begin{tabular}{c}\textbf{Succ. Rate}\\\textbf{(\%)}\end{tabular}}
& \multicolumn{2}{c|}{\begin{tabular}{c}\textbf{Cont. Drift}\\\textbf{(mm)}\end{tabular}}
& \multicolumn{2}{c|}{\begin{tabular}{c}\textbf{Cont. Force}\\\textbf{(N)}\end{tabular}}
& \multicolumn{2}{c|}{\begin{tabular}{c}\textbf{Smooth.-Trans.}\\\textbf{($\mathrm{m/s}^2$)}\end{tabular}}
& \multicolumn{2}{c}{\begin{tabular}{c}\textbf{Smooth.-Rot.}\\\textbf{($\mathrm{rad/s}^2$)}\end{tabular}} \\
\cmidrule(lr){3-4}\cmidrule(lr){5-6}\cmidrule(lr){7-8}\cmidrule(lr){9-10}\cmidrule(lr){11-12}
\multicolumn{2}{c}{\textbf{Admi. (Ours) $\rightarrow$}} 
& w/o & w/ 
& w/o & w/ 
& w/o & w/ 
& w/o & w/ 
& w/o & w/ \\
\midrule
\multirow{3}{*}{ViT-FMT}
& T1  & \textbf{100} & \textbf{100} & 18.12 & \textbf{15.67} & 4.77 & \textbf{5.50} & 3.14 & \textbf{2.98} & 20.29 & \textbf{18.56} \\
& T2  & \textbf{100} & \textbf{100} & 21.04 & \textbf{19.44} & \textbf{0.52} & 0.13 & \textbf{1.87} & 1.95 & \textbf{12.63} & 13.61 \\
& T3  & 80 & \textbf{92} & 25.00 & \textbf{22.08} & 0.96 & \textbf{1.61} & 1.85 & \textbf{1.29} & 14.05 & \textbf{10.38} \\
\midrule
\multirow{3}{*}{ViT-DiT}
& T1 & \textbf{52} & \textbf{52} & 21.79 & \textbf{18.00} & 2.21 & \textbf{2.57} & 2.92 & \textbf{2.54} & 18.73 & \textbf{14.97} \\
& T2 & \textbf{100} & \textbf{100} & 21.22 & \textbf{19.61} & 0.19 & \textbf{0.76} & 4.67 & \textbf{2.41} & 25.93 & \textbf{15.17} \\
& T3 & \textbf{52} & \textbf{52} & 26.22 & \textbf{22.20} & \textbf{1.74} & 0.93 & 2.33 & \textbf{2.05} & 16.47 & \textbf{14.79} \\
\midrule
\multirow{3}{*}{$\pi_{0.5}$}
& T1 & 88 & \textbf{92} & 21.51 & \textbf{11.78} & 2.53 & \textbf{2.80} & \textbf{4.69} & 4.82 & \textbf{29.33} & 29.79 \\
& T2 & 68 & \textbf{76} & 19.78 & \textbf{15.97} & \textbf{0.50} & \textbf{0.50} & 5.06 & \textbf{5.03} & 31.86 & \textbf{31.22} \\
& T3 & \textbf{84} & 76 & 20.18 & \textbf{19.40} & 2.57 & \textbf{3.38} & 4.11 & \textbf{3.53} & 27.26 & \textbf{23.56} \\
\midrule
\multirow{3}{*}{$\Psi_0$}
& T1 & 88 & \textbf{92} & 15.80 & \textbf{13.56} & \textbf{3.14} & 3.03 & 4.20 & \textbf{3.27} & 27.06 & \textbf{23.01} \\
& T2 & 89 & \textbf{96} & 20.48 & \textbf{18.69} & \textbf{2.50} & \textbf{2.50} & \textbf{4.94} & 5.44 & \textbf{32.12} & 33.04 \\
& T3 & 0 & 0 & - & - & - & - & - & - & - & - \\
\bottomrule
\end{tabular}
\end{table*}

Ranked by success rate averaged across tasks \textbf{T1}--\textbf{T3} and both admittance settings (w/o and w/), ViT-FMT performs best (98.7\% on average), followed by $\pi_{0.5}$, while ViT-DiT and the foundation policy $\Psi_0$ trail behind. ViT-DiT often stucks in the initial hugging pose and fails to continue the rotation on the yoga-ball and bucket tasks, whereas $\Psi_0$ produces indecisive motion and fails the bucket task entirely (0\% on \textbf{T3}, leaving its quality metrics undefined).

Across all four backbones, the tactile-based admittance controller regulates the contact centroid and reduces contact drift by approximately $17\%$ on average (Cont.\ Drift columns). Applying admittance control also adjusts the mean contact force as it tracks the predicted force reference (Cont.\ Force columns). Motion quality is on-average improved by the admittance controller: across backbones, translational and rotational accelerations decrease by roughly $14\%$ and $13\%$ (Smooth.-Trans. and Smooth.-Rot. columns). Success rates are mostly unchanged, indicating that the added force feedback improves contact centering without compromising task completion.

The two foundation policies, $\pi_{0.5}$ and $\Psi_0$, are less smooth, as reflected by their higher translational and rotational accelerations. Their jerkier motion can be attributed to the slower inference and more conservative real-time chunking (RTC) settings.

Overall, ViT-FMT is the strongest of the four baseline backbones, so we adopt it as the default backbone in the ablation studies in Sec.~\ref{sec:results_ik_admittance} and Sec.~\ref{sec:result_policy}. Notably, the backbone itself is not a contribution of this work; our contributions, the force-conditioned target-pose correction, the force-supervised planner, and the tactile-based admittance controller, are backbone-agnostic and improve contact quality across all four baselines.

\subsection{Ablation Study of Target-Pose Correction and Admittance Controller on Recorded Data}

In addition to the policy evaluations in Sec.~\ref{sec:results_data} and Sec.~\ref{sec:results_ik_admittance}, we isolate the effects of our target-pose correction and admittance control directly on the recorded data, removing the confounding influence of policy backbones and training setups. We replay the collected yoga-ball trajectories on the robot hardware under four configurations: raw human data, raw teleoperation data, target-pose-corrected human data, and corrected human data with admittance control enabled. This comparison evaluates trajectory feasibility across both data sources. It also tests whether target-pose correction improves the feasibility of human data and whether admittance control improves contact quality. We report the same metrics as in Sec.~\ref{sec:results_data}.

\begin{table}[H]
\centering
\small
\setlength{\tabcolsep}{4pt}
\renewcommand{\arraystretch}{0.92}
\caption{Force-modulation module ablation on data replay across four configurations. ``Failed'' indicates the configuration cannot complete the task. Best value in bold for success rate, contact drift, and motion smoothness.}
\label{tab:module_effectiveness}
\begin{tabular}{l|ccccc}
\toprule
\textbf{Configuration}
& \makecell{\textbf{Success}\\\textbf{Rate (\%)}}
& \makecell{\textbf{Contact Center}\\\textbf{Drift (mm)}}
& \makecell{\textbf{Mean Contact}\\\textbf{Force (N)}}
& \makecell{\textbf{Smooth.-Trans.}\\\textbf{($\text{m/s}^2$)}}
& \makecell{\textbf{Smooth.-Rot.}\\\textbf{($\text{rad/s}^2$)}} \\
\midrule
Raw Human     & Failed         & --             & --            & --            & --             \\
Raw Teleoperation   & 85.35          & 15.59          & \textbf{3.30} & 3.93          & 26.42          \\
Correction      & 89.29          & 17.88          & 3.23          & \textbf{3.57} & \textbf{25.51} \\
Correction + Admittance & \textbf{96.15} & \textbf{11.46} & 3.02          & 4.06          & 27.15          \\
\bottomrule
\end{tabular}
\end{table}

As shown in Table~\ref{tab:module_effectiveness}, \textbf{Raw Human} data fails to complete the task due to loss of contact, confirming that human motion is not directly robot-executable for lack of action labels. In contrast, \textbf{Raw Teleoperation} is executable and completes the task at an $85.35\%$ success rate. Our proposed target-pose \textbf{Correction} improves the feasibility of the human trajectories and raises the success rate to $89.29\%$. Adding admittance control (\textbf{Correction + Admittance}) further raises the success rate to $96.15\%$ and reduces contact center drift by $35.9\%$. This gain comes with a moderate loss of motion smoothness, which is expected because admittance control introduces reactive adjustments based on contact feedback. This ablation confirms that target-pose correction is the key component enabling the feasibility of human data, while admittance control improves contact quality.

\subsection{Admittance Controller Details}

The corrective SE(3) increment $\Delta T^{s}_t$ introduced in Sec.~\ref{sec:lowlevel} contains a rotation $\Delta R^{s}_{xy,t}$ and a translation $\Delta\mathbf{p}^{s}_{z,t}$ derived from force feedback. Both follow proportional control laws,
\begin{equation}
\Delta R^{s}_{xy,t}=\operatorname{Exp}\!\big([(K_R(\mathbf{c}^{s}_{t}-\mathbf{c}^{s,\mathrm{m}}_{t}));\,0]_{\times}\big), \qquad \Delta \mathbf{p}^{s}_{z,t}=K_F(f^{s}_{t}-f^{s,\mathrm{m}}_{t})\,[0,0,1]^\top,
\end{equation}
where $\mathbf{c}^{s}_{t}, \mathbf{c}^{s,\mathrm{m}}_{t} \in \mathbb{R}^2$ are the desired and measured contact centroids in the sensor frame, $f^{s}_{t}$ and $f^{s,\mathrm{m}}_{t}$ are the reference (planner-predicted) and measured normal forces, $K_R\in\mathbb{R}^{2\times2}$ is the contact-centering gain matrix, and $K_F$ is a scalar normal-force gain. The rotation term reorients the palm to re-center the contact centroid, while the translation term drives the local sensor-normal motion to track the reference normal force.

\begin{wrapfigure}[13]{r}{0.5\textwidth}
  \begin{center}
  \vspace{-15pt} % Moves only the image up inside its box
    \includegraphics[width=0.5\textwidth]{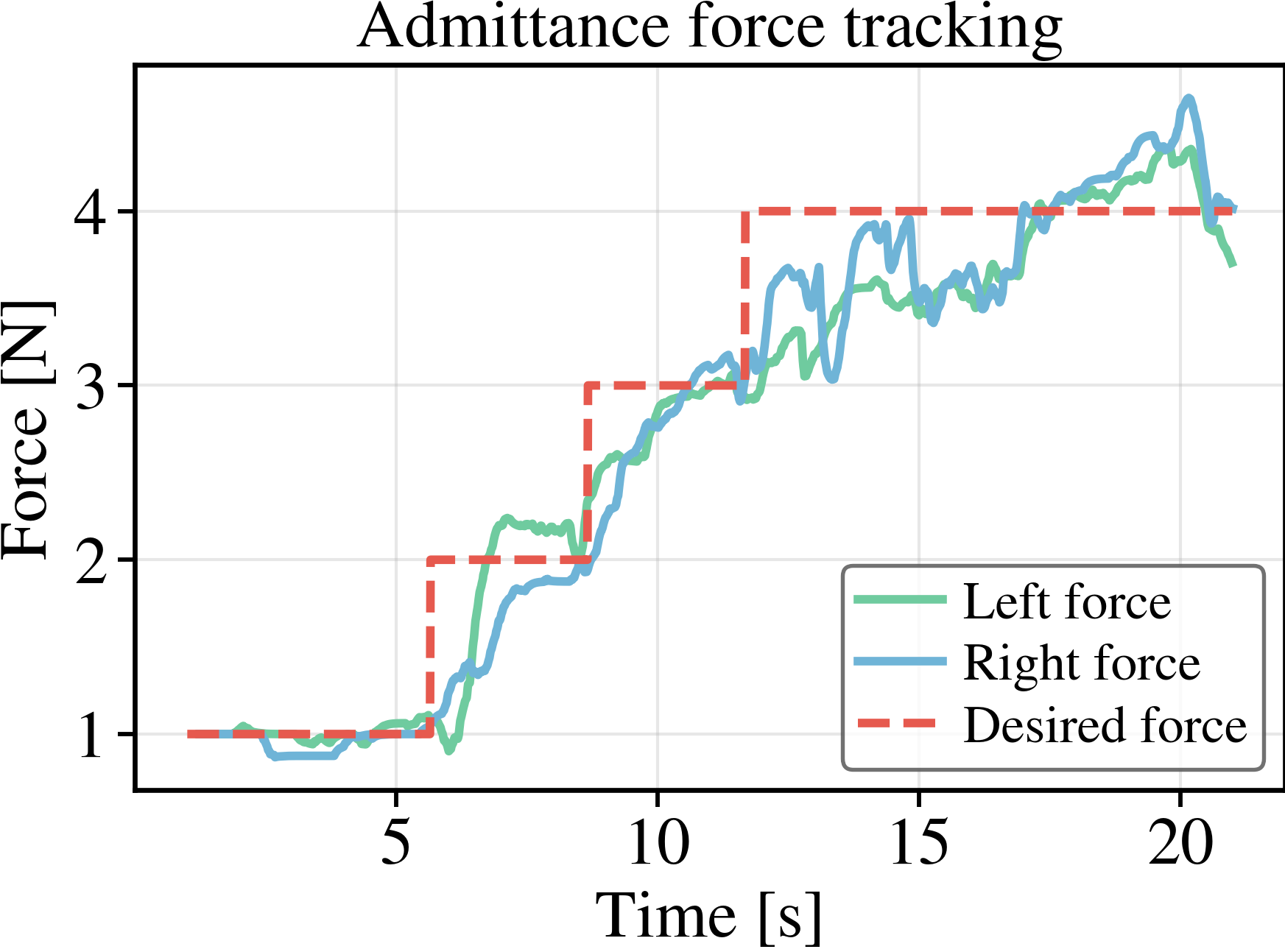}
  \end{center}
  \vspace{-0.2in} % Reduces space between image and caption
  \caption{Admittance controller force-tracking.}
    \label{fig:admittance_tracking}
\end{wrapfigure}
\paragraph{Force regulation evaluation.}
We evaluate the closed-loop force regulation of the tactile admittance controller from Sec.~\ref{sec:lowlevel}. The robot holds a yoga ball between its chest and palms under a fixed pose target, with both palms initially contacting the ball at $1$\,N. It then tracks step changes in the desired contact force $f^{s}_{t}$ from $1$\,N to $4$\,N.

Fig.~\ref{fig:admittance_tracking} shows the force-tracking response: both palms reach each setpoint with steady-state errors within $3.5\,\%$, confirming that the proportional admittance law tracks the planner-predicted force reference accurately. 
%\andy{I dont like this position but its not clipping the page anymore and can probably be dealt with once more text is added above}
\end{document}